\documentclass[journal]{IEEEtran}
\ifCLASSOPTIONcompsoc
  \usepackage[nocompress]{cite}
\else
  \usepackage{cite}
\fi

\ifCLASSINFOpdf
 \else
  \fi

\usepackage{multirow}
\usepackage{tabls}
\usepackage{wrapfig}
\usepackage{hyperref}
\usepackage{amsmath,amssymb} % define this before the line numbering.
\usepackage{amsmath,amssymb}
 \usepackage[ruled,vlined,linesnumbered,resetcount]{algorithm2e}
  \usepackage{booktabs}
 \usepackage[linesnumbered]{algorithm2e}
\usepackage{algorithmic}
\usepackage{caption}
\usepackage{subfigure}
\renewcommand{\algorithmicrequire}{ \textbf{Input:}} %Use Input in the format of Algorithm  
\renewcommand{\algorithmicensure}{ \textbf{Output:}} %UseOutput in
\usepackage[font=small,labelfont=bf]{caption}
   \usepackage{graphicx}
 \usepackage{booktabs}

\usepackage{epstopdf}
\usepackage{graphicx}
\usepackage{subfigure}
\usepackage{enumitem}

\ifCLASSINFOpdf
\else

\fi

\begin{document}
\title{Image De-raining Using a Conditional Generative Adversarial Network}

\author{He Zhang,~\IEEEmembership{Member, IEEE}, Vishwanath Sindagi,~\IEEEmembership{Student Member, IEEE}\\  
        Vishal M. Patel,~\IEEEmembership{Senior Member,~IEEE}% <-this % stops a space
	\thanks{He Zhang is  with Adobe, San Jose, CA email: he.zhang92@rutgers.edu}% <-this % stops a space
	% \thanks{He Zhang is with the department of Electrical and Computer Engineering at Rutgers University, 
	%Piscataway, NJ USA. email: he.zhang92@rutgers.edu. Don't forget to add your email here}
	\thanks{Vishwanath Sindagi and Vishal M. Patel are with the Department of Electrical and Computer Engineering, Johns Hopkins University, Baltimore, MD, USA. Email: \{vsindag1, vpatel36\}@jhu.edu}}

\IEEEtitleabstractindextext{%

\begin{abstract}
Severe weather conditions such as rain and snow adversely affect the visual quality of images captured under such conditions thus rendering them useless for further usage and sharing. In addition, such degraded images drastically affect performance of vision systems. Hence, it is important to address the problem of single image de-raining. However, the inherent ill-posed nature of the problem presents several challenges. We attempt to leverage powerful generative modeling capabilities of the recently introduced Conditional Generative Adversarial Networks (CGAN) by enforcing an additional constraint that the de-rained image must be indistinguishable from its corresponding ground truth clean image. The adversarial loss from GAN provides additional regularization and helps to achieve superior results. In addition to presenting a new approach to de-rain images, we introduce a new refined loss function and architectural novelties in the generator-discriminator pair for achieving improved results. The loss function is aimed at reducing artifacts introduced by GANs and ensure better visual quality. The generator sub-network is constructed using the recently introduced densely connected networks, whereas the discriminator is designed to leverage global and local information to decide if an image is real/fake.  Based on this, we propose a novel single image de-raining method called Image De-raining Conditional Generative Adversarial Network (ID-CGAN), which considers quantitative, visual and also discriminative performance into the objective function. Experiments evaluated on synthetic and real images show that the proposed method outperforms many recent state-of-the-art single image de-raining methods in terms of quantitative and visual performance. Furthermore, experimental results evaluated on object detection datasets using Faster-RCNN also demonstrate the effectiveness of proposed method in improving the detection performance on images degraded by rain. 

\end{abstract}

}

\maketitle

\IEEEdisplaynontitleabstractindextext

\section{Introduction}
\label{sec:introduction}
\begin{figure}[t]
	\centering
	\begin{minipage}{.23\textwidth}
		\centering
		\includegraphics[width=4.2cm,height=3.5cm]{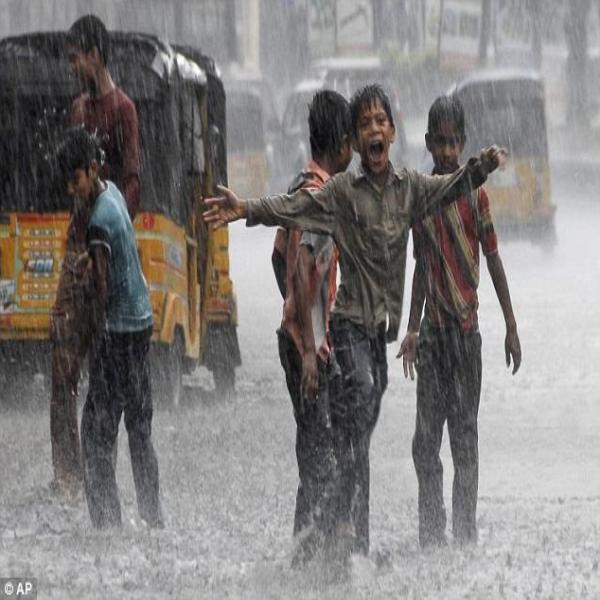}
		\captionsetup{labelformat=empty}
		%\captionsetup{justification=centering}
		%\caption*{(a)}
		%\vspace{1pt}
		%\vskip+6pt
	\end{minipage}
	\begin{minipage}{.23\textwidth}
		\centering
		\includegraphics[width=4.2cm,height=3.5cm]{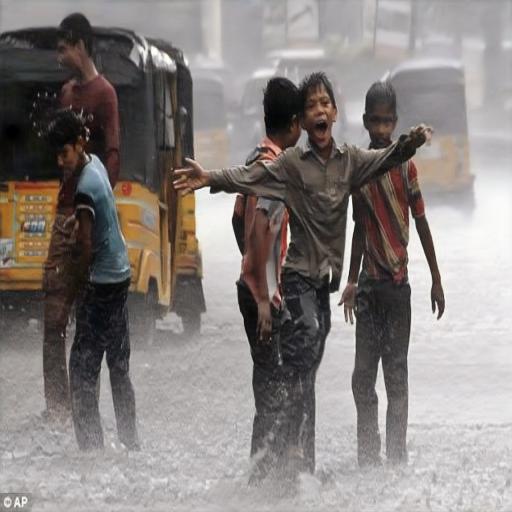}
		\captionsetup{labelformat=empty}
		%\captionsetup{justification=centering}
    	%\caption*{(b)}
		%\vpace{1pt}
		%\vskip+6pt
	\end{minipage}
	\begin{minipage}{.23\textwidth}
		\centering
		\vspace{5pt}
		\includegraphics[width=4.2cm,height=3.2cm]{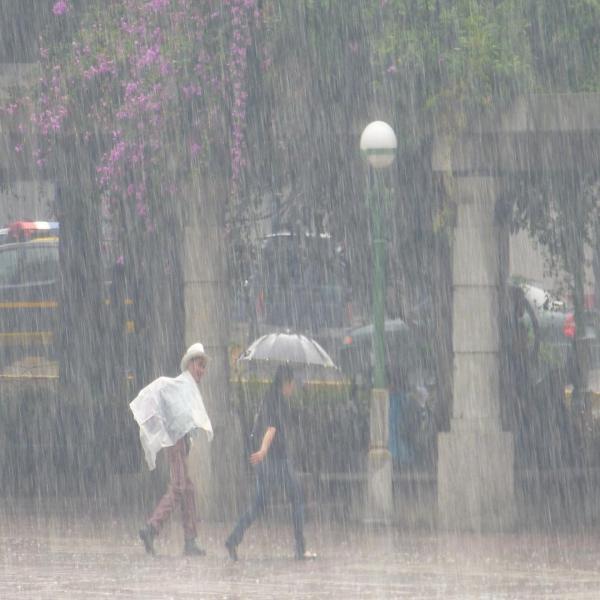}
		\captionsetup{labelformat=empty}
		\captionsetup{justification=centering}
		\caption*{Input }
		%\vskip+6pt
	\end{minipage}	 
	\begin{minipage}{.23\textwidth}
		\centering
		\vspace{5pt}
		\includegraphics[width=4.2cm,height=3.2cm]{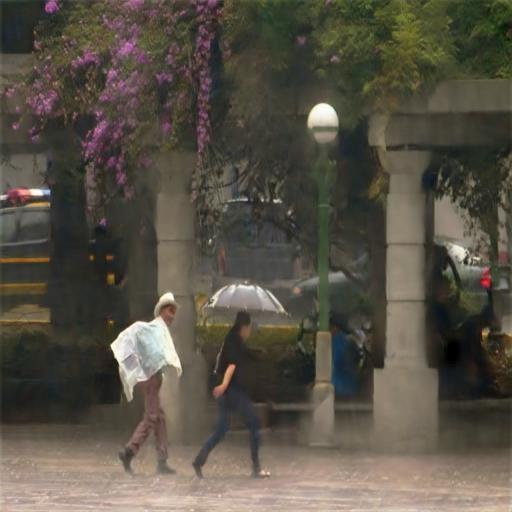}
		\captionsetup{labelformat=empty}
		\captionsetup{justification=centering}
		\caption*{De-rained results}
		%\vskip+6pt
	\end{minipage}	
	\caption{Sample results of the proposed ID-CGAN method for single image de-raining.} \label{fig:over}
\end{figure}

It has been known  that unpredictable impairments such as illumination, noise and severe weather conditions, such as rain, snow and haze,  adversely influence the performance of many computer vision algorithms such as  tracking, detection and segmentation. This is primarily due to the fact that most of  these state-of-the-art algorithms are trained using images that are captured under well-controlled conditions. For example, it can be observed from Fig.~\ref{fig:over}, that the presence of heavy rain  greatly degrade perceptual  quality of the image, thus imposing larger challenge for  face detection and verification algorithms in such weather conditions. A possible method to address this issue is to include images captured under unconstrained conditions during the training process of these  algorithms. However, it may not be practical to collect such large scale datasets as  training set. In addition, in this age of ubiquitous cellphone usage, images captured in the bad weather conditions by cellphone cameras  undergo degradations that drastically affect the visual quality of images making the images useless for sharing and usage. In order to improve the overall quality of such degraded images for better visual appeal and to ensure enhanced performance of vision algorithms, it becomes essential to automatically remove these undesirable artifacts  due to severe weather conditions discussed above. In this paper, we investigate the effectiveness of conditional generative adversarial networks (GANs) in addressing this issue, where a learned discriminator network is used as a guidance to synthesize images free from weather-based degradations. Specifically, we propose a single image-based de-raining algorithm using a conditional GAN framework for visually enhancing images that have undergone degradations due to rain.

\begin{figure}[t!]
	\centering
	\begin{minipage}{.15\textwidth}
		\centering
		\includegraphics[width=2.8cm,height=2.2cm]{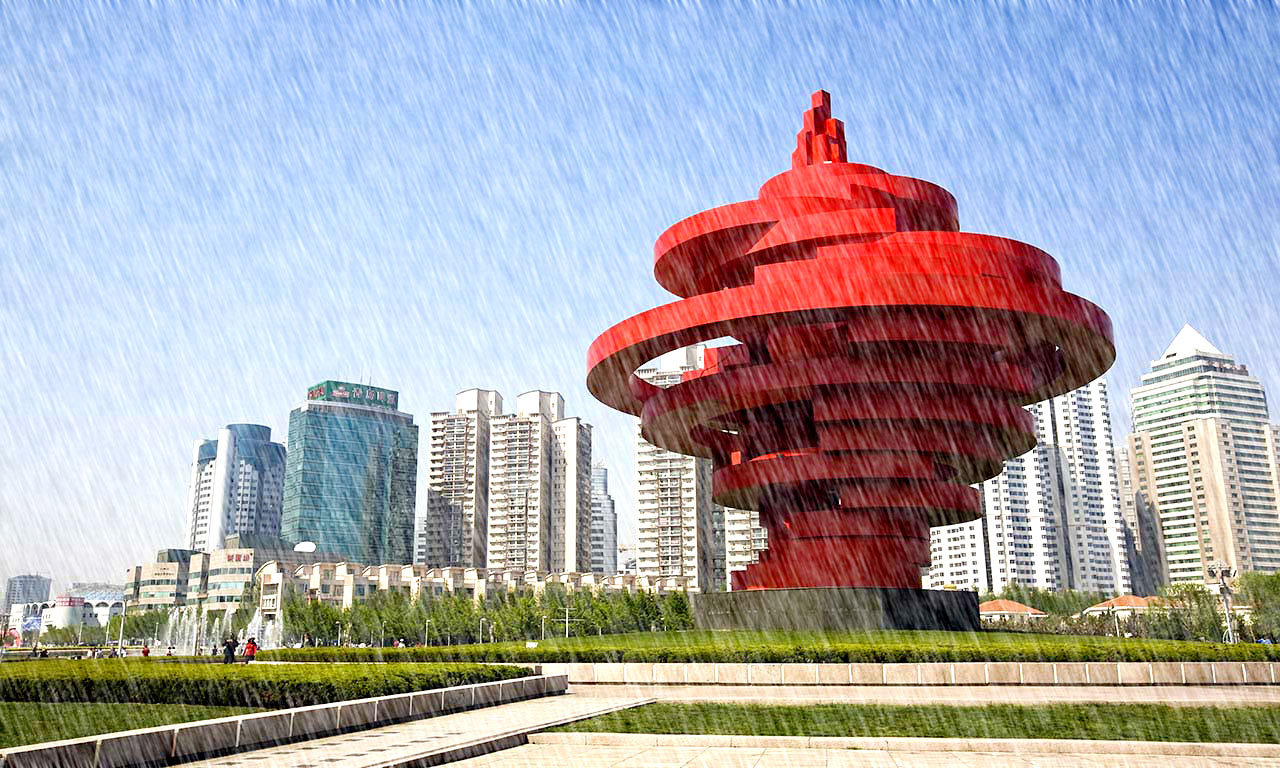}
		\captionsetup{labelformat=empty}
		\captionsetup{justification=centering}
		\caption*{(a)}
		%\vskip+6pt
	\end{minipage}
	\begin{minipage}{.15\textwidth}
		\centering
		\includegraphics[width=2.8cm,height=2.2cm]{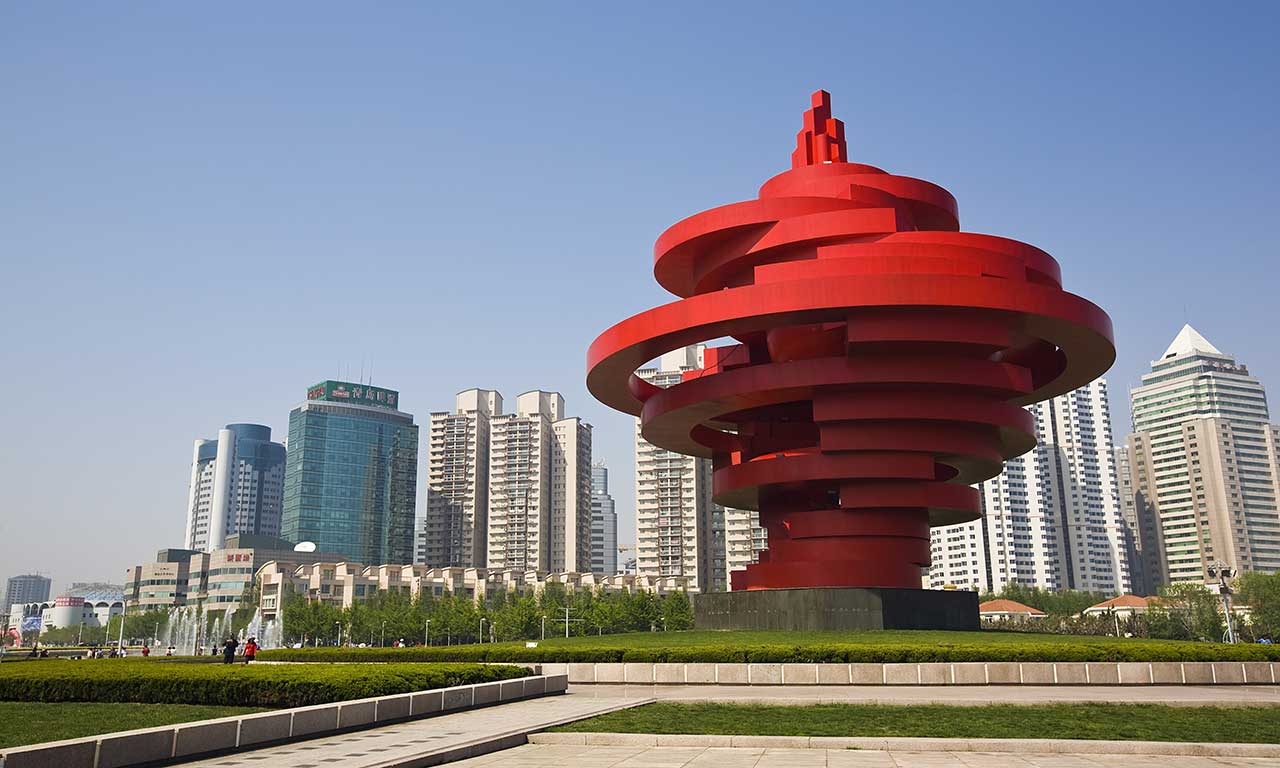}
		\captionsetup{labelformat=empty}
		\captionsetup{justification=centering}
		\caption*{(b)}
		%\vskip+6pt
	\end{minipage}
	\begin{minipage}{.15\textwidth}
		\centering
		\includegraphics[width=2.8cm,height=2.2cm]{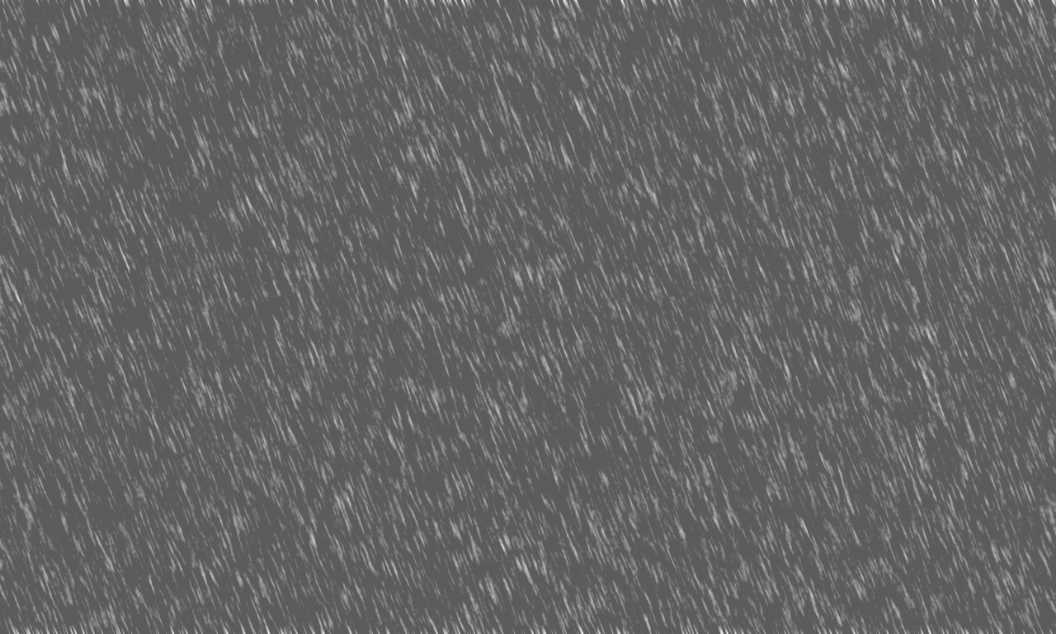}
		\captionsetup{labelformat=empty}
		\captionsetup{justification=centering}
		\caption*{(c)}
		%\vskip+6pt
	\end{minipage}	 
	\caption{Rain streak removal from a single image.  A rainy image (a) can be viewed as the superposition of a clean background image (b) and a rain streak image (c).} \label{fig:over1}
\end{figure}

Mathematically, a rainy image can be decomposed into two separate images: one corresponding to  rain streaks and the other corresponding to the clean background image (see  Fig. ~\ref{fig:over1}).    Hence, the input rainy image can be expressed as \begin{equation}\label{eq:observation}
\mathbf{x}=\mathbf{y}+\mathbf{w},
\end{equation}
where $\mathbf{y}$ represents the clean background image and  $\mathbf{w}$ represents the rain streaks.  As a result, similar to image de-noising  and image separation problems \cite{image_res_cnn,image_separation_BMVC2016,mca_image,mca_diversity,yang2017textured}, image de-raining can be viewed as the problem of separating two components from a rainy image.

%Similar to image de-noising  and image separation \cite{image_res_cnn,image_separation_BMVC2016,mca_image,mca_diversity}, image de-raining can be viewed as the problem of separating two components from a rainy image.  These components are clear background image and rain-streak image.  
%
%modeled as a superposition of two components: clear background image and rain-streak components as follows
%\begin{equation}\label{eq:observation}
%\mathbf{x}=\mathbf{y}+\mathbf{w},
%\end{equation}
%where $\mathbf{y}$ represents the clear background image and  $\mathbf{w}$ represents the rain streaks (see  Fig. ~\ref{fig:over1}). 

%(as illustrated in  Fig. ~\ref{fig:over1}). However, this is an ill-posed problem, as for a given input rainy image $\mathbf{x}$, multiple solutions pairs \{$\mathbf{y}$, $\mathbf{w}$\}  can  be obtained by solving the following equation: 
%\begin{equation}\label{eq:observation}
%\mathbf{x}=\mathbf{y}+\mathbf{w}.
%\end{equation}
%where $\mathbf{y}$ represent clear background image and  $\mathbf{w}$ represent rain-component image. 

In the case of video-based de-raining, a common strategy to solve \eqref{eq:observation} is to leverage additional temporal information, such as methods proposed in \cite{rain_video1,rain_video3,rain_video_2,derain_iccv17_video}. However, temporal information is not available for cases of single image de-raining problem. In such cases,  previous works  have designed  appropriate  prior in solving \eqref{eq:observation}   such as sparsity prior \cite{derain_tip12,dis_rain_2015,derain_tip14,derain_iccv17}, Gaussian Mixture Model (GMM) prior \cite{rain_2016_gmm} and patch-rank prior \cite{derain_lowrank}.   Most recently,  with the large scale synthesized training samples released by de-raining researchers, Convolutional Neural Networks (CNNs)  have been also successfully applied to solve single image de-raining problem \cite{derain_cvpr2017,derain_tip17,derain_cvpr2017_multi, zhang2018density}. By directly  learning a non-linear mapping between the input rainy image and its corresponding ground truth using a CNN structure with some prior information enforced, CNN-based methods are able to achieve superior visual performance. 

Even though tremendous improvements have been achieved, we note that these methods do not consider   additional information into the optimization.  Hence, to design a visually appealing de-raining algorithm, we must consider the following information into the optimization framework:
\begin{enumerate}
\item[(a)] The criterion that performance of vision algorithms such as detection and classification should not be affected by the presence of rain streaks should be considered in the objective function. The inclusion of this discriminative information ensures that the reconstructed image is indistinguishable from its original counterpart. 
\item[(b)] Rather than concentrating only on the characterization of rain-streaks, visual quality may also be considered into the optimization function. This can  ensure that the de-rained image looks visually appealing without losing important details. 
\item[(c)] Some of the existing methods adopt off-line additional image processing techniques  to enhance the results \cite{derain_tip17,derain_tip12}.  Instead, it would be better to use a more unified structure to deal with the problem without any additional processing.  
\end{enumerate}

In this work, these criteria are incorporated in a novel conditional GAN-based framework called Image De-raining Conditional Generative Adversarial Network (ID-CGAN) to address the single image de-raining problem. We aim to leverage the generative modeling capabilities of the recently introduced CGANs. While existing CNN-based approaches minimize only L2 error, these methods need additional regularization due to the ill-posed nature of the problem. In this work, adversarial loss from CGANs is used as additional regularizer leading to superior results in terms of visual quality and quantitative performance. The use of discriminator for classifying between real/fake samples provides additional feedback, enabling the generator to produce results that are visually similar to the ground-truth clean samples (real samples). Inspired by the recent success of GANs for pixel-level vision tasks such as image generation \cite{GAN_unsupervised}, image inpainting \cite{Gan_context} and image super-resolution \cite{GAN_sisr}, our network consists of two sub-networks: 	densely-connected generator (G) and multi-scale discriminator (D). The generator acts as a mapping function to translate an input rainy image to de-rained image such that it fools the discriminator, which is trained to distinguish rainy images from images without rain. The discriminator is designed to capture hierarchical context information through multi-scale pooling. However, traditional GANs \cite{GAN_unsupervised} are not stable to train and  may introduce artifacts in the output image  making it visually unpleasant and artificial.  To address this issue, we introduce a new refined perceptual loss to serve as an additional loss function to aid the proposed network in generating visually pleasing outputs. Furthermore, to leverage different scale information in determining whether the corresponding de-rained image is real or fake, a multi-scale discriminator is proposed.  Sample results of the proposed ID-CGAN algorithm are shown in Fig.~\ref{fig:over}.  In summary, this paper makes the following contributions:
\begin{enumerate}
\item A conditional GAN-based framework to address the challenging single image de-raining problem without the use of any additional post-processing.
\item A densely-connected generator sub-network that is specifically designed for the single image de-raining task. 
\item A multi-scale discriminator is  proposed  to 
leverage both local and global information to determine whether the corresponding de-rained image is real or fake.
\item Extensive experiments  are conducted on publicly available and synthesized datasets to demonstrate the effectiveness of the proposed method in terms of visual quality and quantitative performance. 
/%Detailed qualitative and quantitative comparisons with existing state-of-the-art methods are presented  \footnote{Datasets and experimental implementation are available at\\ \url{http://www.rci.rutgers.edu/~vmp93/index_ImageDeRaining.html}}.
\item Lastly, effectiveness of the proposed method in improving high-level object detection task is demonstrated on VOC dataset \cite{pascal}. The detections are performed using Faster-RCNN \cite{faster_rcnn}.

\end{enumerate}

This paper is organized as follows.   A brief background on de-raining, GANs and perceptual loss is given in Section~\ref{sec:related}.   The details of the proposed ID-CGAN method are given in Section~\ref{sec:method}.  Experimental results on both synthetic and real images are presented in Section~\ref{sec:exp}.  Finally, Section~\ref{sec:con} concludes the paper with a brief summary and discussion.

\section{Background}\label{sec:related}
In this section, we briefly review the literature for existing single image de-raining methods and conditional GANs.   
\subsection{Single Image De-raining}
As discussed in Section \ref{sec:introduction}, single image de-raining  is an extremely challenging task due to its ill-posed nature. In addition, the  unavailability of temporal information, which could have been used as additional constraints, also pose challenges to solve the single image de-raining problem. Hence, in order to generate optimal solutions to this problem, different kinds of prior knowledge are enforced into the optimization framework.  In the following, we discuss sparsity-based methods, low-rank method, gaussian mixture model methods and deep learning methods in solving image de-raining problem.

\noindent\textbf{Sparsity-based Methods:}
To overcome the issue of ill-posed nature of  \eqref{eq:observation}, authors in \cite{derain_tip12} employed two set of learned dictionary atoms (filters) to sparsely represent  clean background image and rain-streak separately. The separation is performed in the high frequency part of the given rainy  image, assuming that the low-frequencies correspond to the clean background image.  An important assumption that is made in this approach is that rain streaks in an image usually have similar orientations. Similarly, Luo \textit{et al.} in \cite{dis_rain_2015}  proposed a discriminative approach that sparsely
approximates the patches of clean background and rain-streak components  by discriminative codes over two set of  learned dictionary atoms with strong mutual exclusivity property. However, their method generates artifacts around the rain-streak components in the resulting images. 

\noindent\textbf{Low-rank Representation-based Methods}
Inspired by the observation that rain streak components within an  image share similar patterns and  orientations, Chen \textit{et al.} proposed a low patch-rank  prior to capture these  patterns. This is motivated by the use of patch-rank for characterizing the texture component in solving cartoon-texture image decomposition problem \cite{BNN_rank,lowrank_path}.  Since the patch-rank may also capture some repetitive texture patterns, this method    removes important texture details from the input image, due to which the results become blurred.  To address this issue, Zhang \textit{et al.} recently proposed a convolutional coding-based method \cite{rain_wacv2017} that uses a set of  learned convolutional low-rank filters to capture the rain pixels.

\noindent\textbf{Gaussian Mixture Model-based Methods:}
Based on the assumptions that GMMs can accommodate multiple orientations and scales of rain streaks,  Li \textit{et al.} in \cite{rain_2016_gmm} used the image decomposition framework to propose patch-based GMM priors to model background and rain streaks separately.

\noindent\textbf{Deep Learning-based Methods}

The success of convolutional neural networks in several computer vision tasks  \cite{deep_res,deep_supervision,dense_net,vgg} has inspired researchers to develop CNN-based approaches for image de-raining \cite{derain_cvpr2017,derain_tip17,derain_cvpr2017_multi}. These methods attempt to learn a non-linear function to convert an input rainy image to a clean target image.  Based on  the observation that both rain streaks and object details remain only in the detail layer, Fu \emph{et al.} \cite{derain_tip17} employed a two-step procedure, where the input rainy image is decomposed into a background-based layer and a detail layer separately. Then, a CNN-based non-linear mapping is learned to remove the rain streaks from the detail layer. Built on  \cite{derain_tip17}, Fu \emph{et al.} extended the network  structure  using Res-block \cite{deep_res} in \cite{derain_cvpr2017}. Yang \emph{et al.} proposed a CNN  structure that can jointly detect and  remove rain streaks. Most recently, several deep learning methods have been explored for single image de-raining task \cite{taixiang_derain_cnn,wei2018semi,fu_derain_light,fan_derain_acmmm}.

\begin{table*}[htp!]
	\centering
	\label{my-label}
	\resizebox{\textwidth}{!}{%
\begin{tabular}{|c|c|c|c|c|c|c|}
	\hline
	& \begin{tabular}[c]{@{}c@{}}No addition pre-  \\ (or post) processing\end{tabular} & End-to-end mapping & \begin{tabular}[c]{@{}c@{}}Consider discriminative performance \\ in the optimization\end{tabular} & \begin{tabular}[c]{@{}c@{}}Consider  visual  performance \\ in the optimization\end{tabular} & Not Patch-based & Time efficiency \\ \hline\hline
	SPM \cite{derain_tip12} &  &  &  &  &  &  \\ \hline
	PRM \cite{derain_lowrank}  & $\surd$ &  &  &  & &  \\ \hline
	DSC \cite{dis_rain_2015} &$\surd$  &  &  &  &  &  \\ \hline
	CNN \cite{derain_tip17} &  &$\surd$  &  &  &$\surd$  & $\surd$ \\ \hline
	GMM \cite{rain_2016_gmm} & $\surd$ &  &  &  &$\surd$  & $\surd$ \\ \hline
	CCR \cite{rain_wacv2017} & $\surd$  &  &  &  & $\surd$ &  \\ \hline
	DDN \cite{derain_cvpr2017} &  & $\surd$  &  &  & $\surd$ & $\surd$ \\ \hline
	JORDER \cite{derain_cvpr2017_multi} & $\surd$  & $\surd$  &  &  & $\surd$ & $\surd$  \\ \hline
	ID-CGAN  & $\surd$  & $\surd$  & $\surd$ & $\surd$ & $\surd$ & $\surd$ \\ \hline
\end{tabular}%
	}
\caption{Compared to the existing methods, our ID-CGAN  has several desirable properties: 1. No additional image processing. 2. Include discriminative factor into optimization. 3. Consider visual performance into optimization.}\label{tbl:methods}
\end{table*}

\subsection{Generative Adversarial Networks }
Generative Adversarial Networks \cite{gan}(GANs) are a class of methods to model  a data distributions  and consist of  two functions: the generator $G$, which translate a sample  from a random uniform  distribution to the data distribution; the discriminator $D$ which measure the probability whether a given  sample  belongs to the data distribution or not. Based on a game theoretic min-max principles, the generator and discriminator are typically learned jointly by alternating the training of D and G. Although GANs are able to  generate visual appealing  images by preserving high frequency details, yet GANs still face many unsolved challenges: in general they are notoriously difficult to train and GANs easily suffer from modal collapse.  Recently, researchers have explored various aspects of GANs such as leveraging another conditional variable \cite{gan_conditioanl}, training improvements \cite{salimans2016improved} and use of task specific cost function \cite{creswell2016task}. Also, an alternative viewpoint for the discriminator function is explored by Zhao \textit{et al.} \cite{zhao2016energy} where they deviate from the traditional probabilistic interpretation of the discriminator model. 

The success of GANs in synthesizing visually appealing images has inspired researchers to explore the use of  GANs  in other related  such as text-to-image synthesis \cite{zizhao_cvpr_text,xu2017attngan,gan_stack_zhang}, single image super-resolution \cite{GAN_sisr}, domain adaption \cite{liu2016coupled}, face synthesis  \cite{di2019polarimetric} and other related applications \cite{zizhao_cvpr_medical,yizhe_cvpr18,he_2018_CVPR,yu2018generative,xi_peng_gan,xing_icpr}.

\begin{figure*}[t!]
	\centering
	\vskip-5pt
	\begin{minipage}{1\textwidth}
		\includegraphics[width=7.2in,height=3.5in]{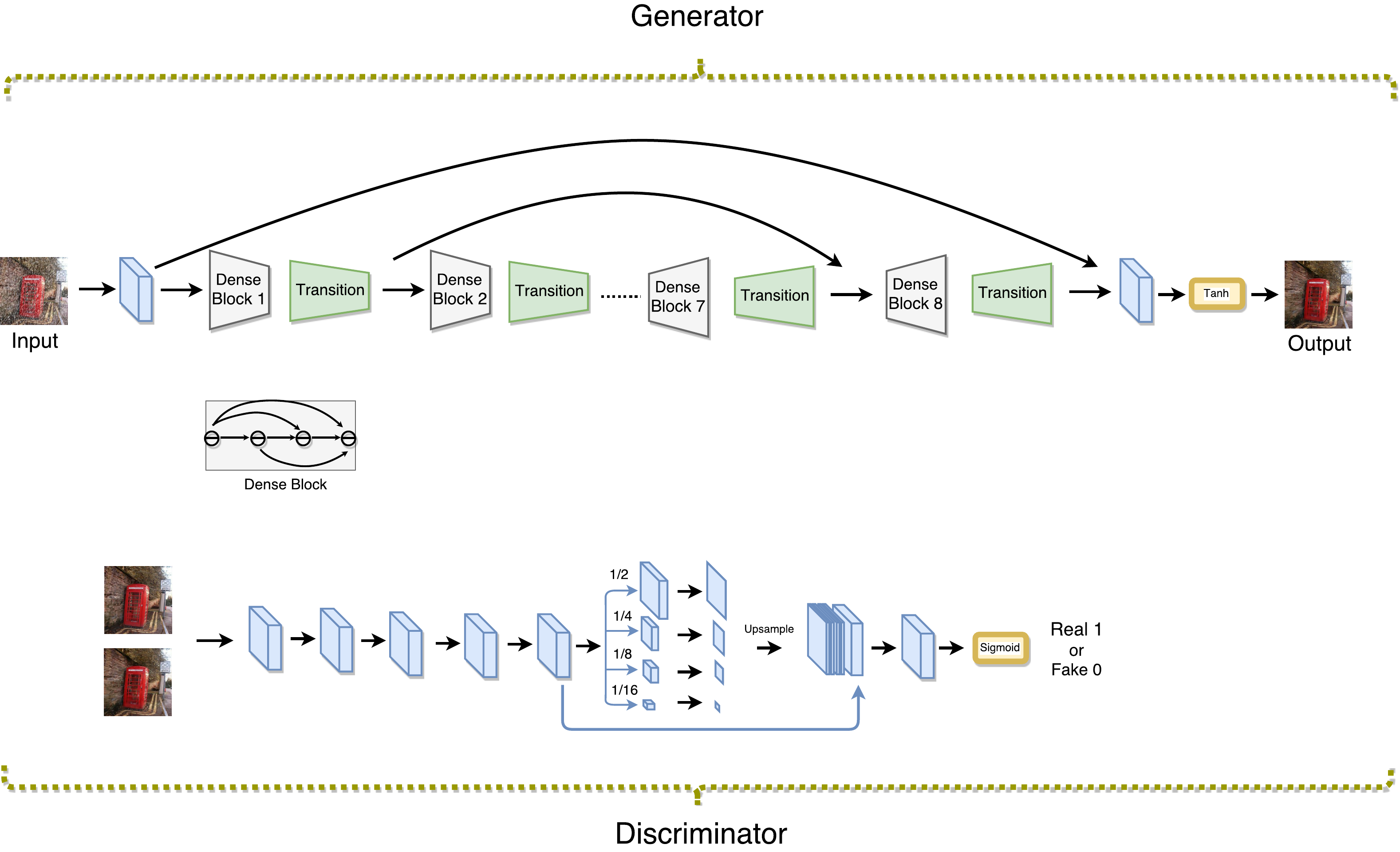}
		\captionsetup{labelformat=empty}
		\captionsetup{justification=centering}
	\end{minipage}
	\caption{An overview of the proposed ID-CGAN method for single image de-raining. The network consists of two sub-networks: generator $G$ and discriminator $D$.} 
	\label{fig:overview}
\end{figure*}

\section{Proposed Method}\label{sec:method}
Instead of solving  \eqref{eq:observation} in a decomposition framework, we aim to directly learn a mapping from an input rainy image to a de-rained (background) image by constructing a conditional GAN-based deep network called ID-CGAN. The proposed network is composed of three important parts  (generator, discriminator and perceptual loss function) that serve distinct purposes. Similar to traditional GANs \cite{pix2pix2016,gan}, the proposed method contains two sub-networks: a generator sub-network ${G}$ and a discriminator sub-network ${D}$. The generator sub-network ${G}$ is a densely-connected symmetric deep CNN network with appropriate skip connections as shown in the top part in Fig.~\ref{fig:overview}. Its primary goal is to synthesize a de-rained image from an image that is degraded by rain (input rainy image). The multi-scale discriminator sub-network $D$, as shown in the bottom part in Fig.~\ref{fig:overview}, serves to distinguish `fake' de-rained image (synthesized by the generator) from corresponding ground truth `real' image. It can also be viewed as a guidance for the generator $G$. Since GANs are known to be unstable to train which results in artifacts in the output image synthesized by $G$, we define a refined perceptual loss functions to address this issue. Additionally, this new refined loss function ensures that the generated (de-rained) images are visually appealing. In what follows, we elaborate these modules in further detail, starting with the GAN objective function followed by details of the network architecture for the generator/discriminator and the overall loss function.

\subsection{GAN Objective Function}

The objective of a generative adversarial network is based on the mini-max game, where the idea is to  learn a generator $G$ that synthesizes samples similar to the data distribution such that the discriminator is not able to distinguish between the synthesize samples and real samples. At the same time, the goal is also to learn a good discriminator $D$ such that it is able to distinguish between synthesized and real samples. To achieve this,  the proposed method alternatively updates $G$ and $D$ following the structure proposed in \cite{gan,pix2pix2016}. Note that in our work, we use a conditional variant of GAN, where the generator $G$, learns to generate a mapping from a condition variable. Given an input rainy image $\mathbf{x}$, conditional GAN learns a non-linear  function to synthesize the  output image $\mathbf{y}$ by conditioning on the input image $\mathbf{x}$:
\begin{equation}\label{eq:GAN1}
\begin{split}
\min_G \max_D \quad & \mathbb E _{\mathbf{x}\sim p_{data(\mathbf{x})}, }[\log (1- D(\mathbf{x}, G(\mathbf{x})))]+\\
&\mathbb E _{\mathbf{x}\sim p_{data(\mathbf{x},\mathbf{y})}}[\log D(\mathbf{x},\mathbf{y}))].
\end{split}
 \end{equation}

\subsection{Generator with Symmetric Structure}
As the goal of single image de-raining is to generate pixel-level de-rained image, the generator should be able to remove rain streaks as much as possible without loosing any detail information of the background image.  So the  key part lies in designing a good structure to generate de-rained image.    

Existing methods for solving \eqref{eq:observation}, such as sparse coding-based methods \cite{derain_tip12,MCA_ATS,mca_image,mca_diversity}, neural network-based methods \cite{denoise_auto} and CNN-based methods \cite{image_res_cnn}  have all adopted a symmetric (encoding-decoding) structure.  For example,  sparse coding-based methods use a learned or pre-defined synthesis dictionaries to decode the input noisy image into sparse coefficient map.  Then another set of analysis dictionaries are used to  transfer the coefficients to desired clear output. Usually, the input rainy image is transferred to a specific domain  for effective separation of background image and undesired component (rain-streak). After separation, the background image (in the new domain) has to be transferred back to the original domain which requires the use of a symmetric process. 

Following these methods, a symmetric structure is  adopted to form the generator sub-network.  The generator $G$ directly learns an end-to-end mapping from input rainy  image to its corresponding  ground truth. In contrast to the existing adversarial networks for image-to-image translation that use U-Net \cite{unet,pix2pix2016} or ResNet blocks \cite{deep_res,SR_photorea} in their generators, we use the recently introduced densely connected blocks \cite{huang2017densely}. These dense blocks enable strong gradient flow and result in improved parameter efficiency. Furthermore, we introduce skip connections across the dense blocks to efficiently leverage features from different levels and guarantee better convergence. The $j$th dense block $\mathbf{D}_j$ is represented as:
\begin{equation}\label{eq:single_stream}
\mathbf{D}_j={cat}[D_{j,1}, D_{j,2},..., D_{j,6}],
\end{equation}
where $D_{j,i}$ represents the features from the $i$th layer in dense block $\mathbf{D}_j$ and each layer in a dense block consists of three consecutive operations, batch normalization (BN), leaky rectified linear units (LReLU) and a 3$\times$3 convolution. 

Each dense block is followed by a transition block ($T$), functioning as up-sampling ($Tu$), down-sampling ($Td$) or no-sampling operation ($Tn$).  To make the network efficient in training and have better convergence performance,  symmetric skip connections are included into the proposed generator sub-network, similar to \cite{image_res_cnn}.   The generator network is as follows:\linebreak
\emph{CBLP(64)-D(256)-Td(128)-D(512)-Td(256)-D(1024)-Tn(512)-D(768)-Tn(128)-D(640)-Tu(120)-D(384)-Tu(64)-D(192)-Tu(64)-D(32)-Tn(16)-C(3)-Tanh}\linebreak
\noindent where, $CBLP$ is a set of convolutional layers followed by batch normalization, leaky ReLU activation and pooling module, and the number inside braces indicates the number of channels for the output feature maps of each block.  Details of the architecture is also shown in Table ~\ref{tab:network}.

\begin{table}[]
	\centering
	\begin{tabular}{|c|}
		\hline
		Architecture \\ \hline \hline
		Input, num\_c=3 \\ \hline
		\begin{tabular}[c]{@{}c@{}}3x3 Convolution, BN, ReLu, MaxP\\ num\_c=64\end{tabular} \\ \hline
		D(4 layers)+Td, num\_c=128 \\ \hline
		D(6 layers)+Td, num\_c=256 \\ \hline
		D(8 layers)+Tn, num\_c=512 \\ \hline
		D(8 layers)+Tn, num\_c=128 \\ \hline
		D(6 layers)+Tu, num\_c=120 \\ \hline
		D(4 layers)+Tu, num\_c=64 \\ \hline
		D(4 layers)+Tu, num\_c=64 \\ \hline
		D(4 layers)+Tn, num\_c=16 \\ \hline
		3x3 Conv, Tanh, num\_c=3 \\ \hline
		Ourput, num\_c=3 \\ \hline
	\end{tabular}
\caption{Network architecture for generator.}
	\label{tab:network}
\end{table}
\subsection{Multi-scale Discriminator}
\begin{figure*}[!]
	\centering
	\begin{minipage}{.135\textwidth}
		\centering
		\includegraphics[width=2.5cm,height=2.2cm]{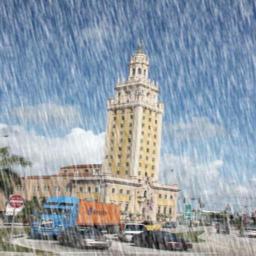}
		\captionsetup{labelformat=empty}
		\captionsetup{justification=centering}
		%\vskip+6pt
	\end{minipage}
	\begin{minipage}{.135\textwidth}
		\centering
		\includegraphics[width=2.5cm,height=2.2cm]{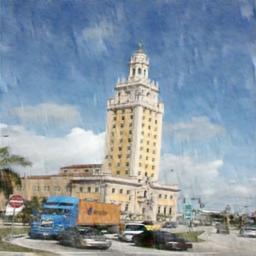}
		\captionsetup{labelformat=empty}
		\captionsetup{justification=centering}
		%\vskip+6pt
	\end{minipage}
	\begin{minipage}{.135\textwidth}
		\centering
		\includegraphics[width=2.5cm,height=2.2cm]{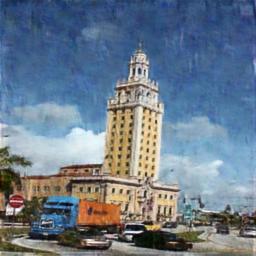}
		\captionsetup{labelformat=empty}
		\captionsetup{justification=centering}
		%\vskip+6pt
	\end{minipage}
	\begin{minipage}{.135\textwidth}
		\centering
		\includegraphics[width=2.5cm,height=2.2cm]{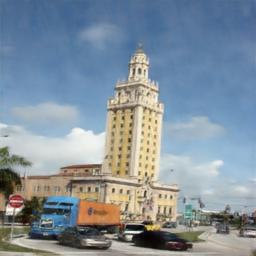}
		\captionsetup{labelformat=empty}
		\captionsetup{justification=centering}
		%\vskip+6pt
	\end{minipage}
	\begin{minipage}{.135\textwidth}
		\centering
		\includegraphics[width=2.5cm,height=2.2cm]{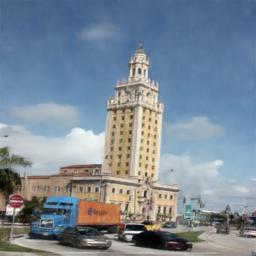}
		\captionsetup{labelformat=empty}
		\captionsetup{justification=centering}
		%\vskip+6pt
	\end{minipage}	 
	\begin{minipage}{.135\textwidth}
		\includegraphics[width=2.5cm,height=2.2cm]{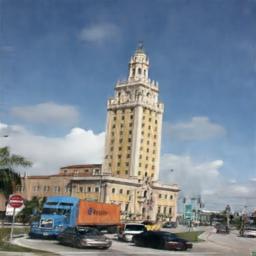}
		\captionsetup{labelformat=empty}
		\captionsetup{justification=centering}
		%\vskip+6pt
	\end{minipage}	
	\begin{minipage}{.135\textwidth}
		\centering
		\includegraphics[width=2.5cm,height=2.2cm]{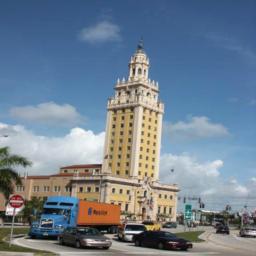}
		\captionsetup{labelformat=empty}
		\captionsetup{justification=centering}
		%\vskip+6pt
	\end{minipage}\\		\vskip+6pt
	\begin{minipage}{.135\textwidth}
		\centering
		\includegraphics[width=2.5cm,height=2.5cm]{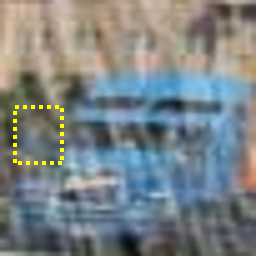}
		\captionsetup{labelformat=empty}
		\captionsetup{justification=centering}
		\caption*{(a)}
	\end{minipage}
	\begin{minipage}{.135\textwidth}
		\centering
		\includegraphics[width=2.5cm,height=2.5cm]{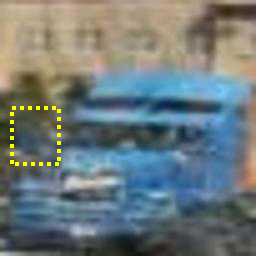}
		\captionsetup{labelformat=empty}
		\captionsetup{justification=centering}
		\caption*{(b)  }
	\end{minipage}
	\begin{minipage}{.135\textwidth}
		\centering
		\includegraphics[width=2.5cm,height=2.5cm]{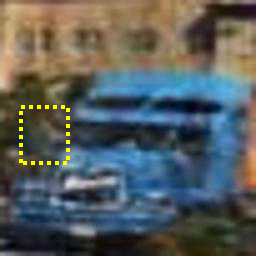}
		\captionsetup{labelformat=empty}
		\captionsetup{justification=centering}
		\caption*{(c)  }
	\end{minipage}
	\begin{minipage}{.135\textwidth}
		\centering
		\includegraphics[width=2.5cm,height=2.5cm]{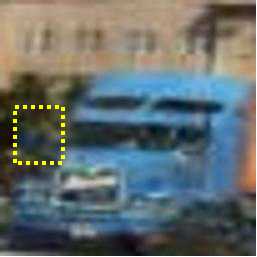}
		\captionsetup{labelformat=empty}
		\captionsetup{justification=centering}
		\caption*{(d)  }
	\end{minipage}
	\begin{minipage}{.135\textwidth}
		\centering
		\includegraphics[width=2.5cm,height=2.5cm]{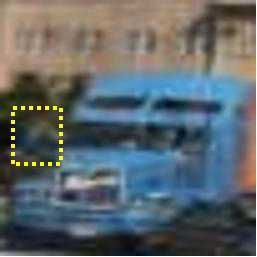}
		\captionsetup{labelformat=empty}
		\captionsetup{justification=centering}
		\caption*{(e)  }
	\end{minipage}	 
	\begin{minipage}{.135\textwidth}
		\includegraphics[width=2.5cm,height=2.5cm]{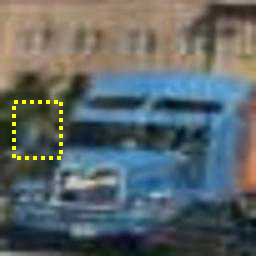}
		\captionsetup{labelformat=empty}
		\captionsetup{justification=centering}
		\caption*{(f)  }
	\end{minipage}
	\begin{minipage}{.135\textwidth}
		\centering
		\includegraphics[width=2.5cm,height=2.5cm]{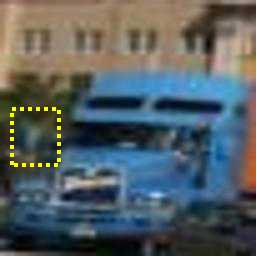}
		\captionsetup{labelformat=empty}
		\captionsetup{justification=centering}
		\caption*{(g)  }
	\end{minipage}
	\caption{Qualitative comparisons for different baseline configurations of the proposed method. (a) Input image, (b) GEN, (c) GEN-CGAN-S, (d) GEN-P, (e) GEN-CGAN-PS, (f) ID-CGAN and (g) Target image.  }
	\label{fig:baseline_real1}
\end{figure*}

From the point of view of a GAN framework, the goal of de-raining an input rainy image is not only to make the de-rained result visually appealing and quantitatively comparable to the ground truth, but also to ensure that the de-rained result is indistinguishable from the ground truth image. Therefore, a learned discriminator sub-network is designed to classify if each input image is real or fake. Previous methods \cite{pix2pix2016} have demonstrated the effectiveness of leveraging an efficient patch-discriminator in generating high quality results. For example, Isola \emph{et al} \cite{pix2pix2016} adopt a 70$\times$ 70 patch discriminator, where $70\times 70$ indicates the receptive field of the discriminator. Though such a single scale (eg. 70$\times$ 70) patch-discriminator is able to achieve visually pleasing results, however, it is still not  capable enough to  capture the  global context information, resulting in insufficient estimation.  As shown in the zoomed-in part of the Fig.~\ref{fig:baseline_real1} (e), it can be observed that certain tiny details are still missing in the de-rained results using a single scale discriminator. For example, it can be observed from the second  row of  Fig.~\ref{fig:baseline_real1} that the front mirror of truck is largely being removed in the  de-rained results. This is probably due to the fact that the receptive field size in the discriminator is 70$\times$ 70 and no additional surrounding context is provided. Hence, we argue that it is important to leverage a more powerful  discriminator that captures both local and global information to decide whether it is real or fake. 

To effectively address this issue, a novel   multi-scale discriminator  is proposed in this paper. This is inspired by the usage of multi-scale features in objection detection \cite{spp} and semantic segmentation \cite{psp_net}. Similar to the structure that was proposed in \cite{pix2pix2016}, a convolutional layer with  batch normalization and PReLU activation are used as a basis throughout the discriminator network. Then,  a multi-scale  pooling module, which pools features at different scales,  is stacked at the end of the discriminator.   The pooled features are then upsampled and concatenated, followed by a 1$\times$1 convolution and a sigmoid function to produce a probability score normalized between [0,1].  By using features at different scales, we explicitly incorporate global hierarchical context into the discriminator. The proposed discriminator sub-network $D$ is shown in the bottom part of Fig. \ref{fig:overview}.  And details of the multi-scale discriminator is shown in Table~\ref{ta:dis}.
\begin{table}[]
	\centering
	\begin{tabular}{|c|}
		\hline
		Architecture \\ \hline \hline
		Input, num\_c=6 \\ \hline
		\begin{tabular}[c]{@{}c@{}}3x3 Convolution, BN, ReLu, MaxP,\\ num\_c=64\end{tabular} \\ \hline
		\begin{tabular}[c]{@{}c@{}}3x3 Convolution, BN, ReLu, MaxP,\\ num\_c=256\end{tabular} \\ \hline
		\begin{tabular}[c]{@{}c@{}}3x3 Convolution, BN, ReLu, MaxP,\\ num\_c=512\end{tabular} \\ \hline
		\begin{tabular}[c]{@{}c@{}}3x3 Convolution, BN, ReLu, MaxP, \\ num\_c=64\end{tabular} \\ \hline
		Four-level Poling Module, num\_c=72 \\ \hline
		Sigmoid \\ \hline
		Output, num\_c=72 \\ \hline
	\end{tabular}
\caption{Network architecture for the discriminator. }
\label{ta:dis}
\end{table}

%The structure of the discriminator sub-network is as follows:\\ \emph{CB($K_2$)-CBP(2$K_2$)-CBP(4$K_2$)-CBP(8$K_2$)-C(1)-Sigmoid} \\

%\noindent where, $CB(K_2)$ is a set of $K_2$ channel convolutional layers followed by batch normalization and $C(1)$ is a set of $1$-channel convolutional layers. 

\begin{figure*}[htp!]
	\vskip+5pt
	\centering
	\includegraphics[width=1.06in]{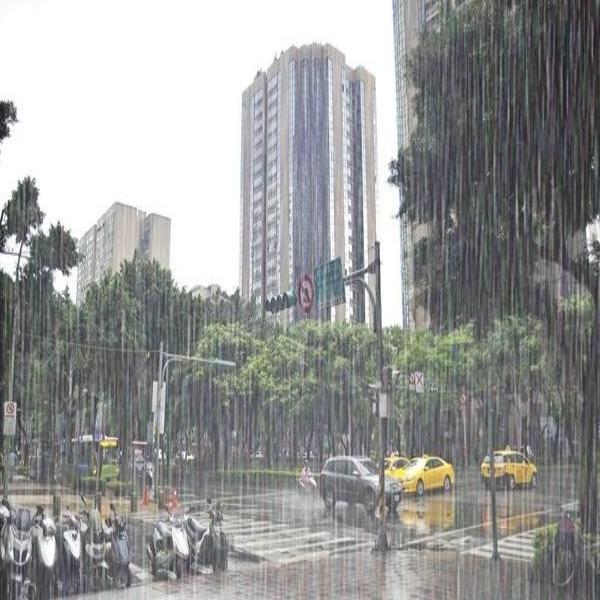}
	\includegraphics[width=1.06in]{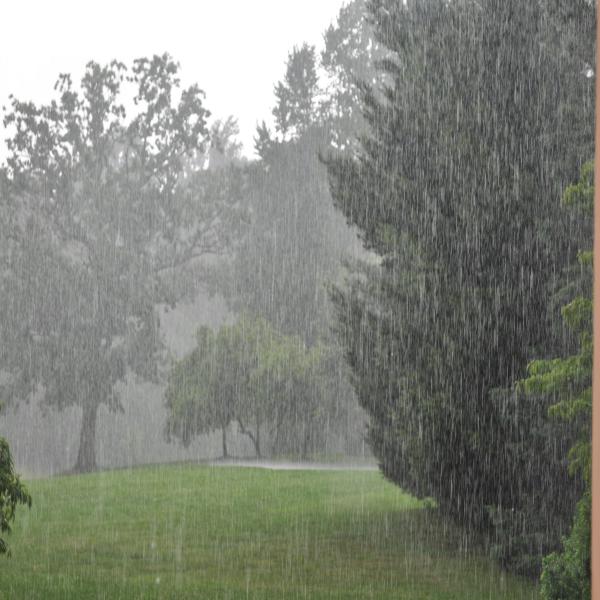}
	\includegraphics[width=1.06in]{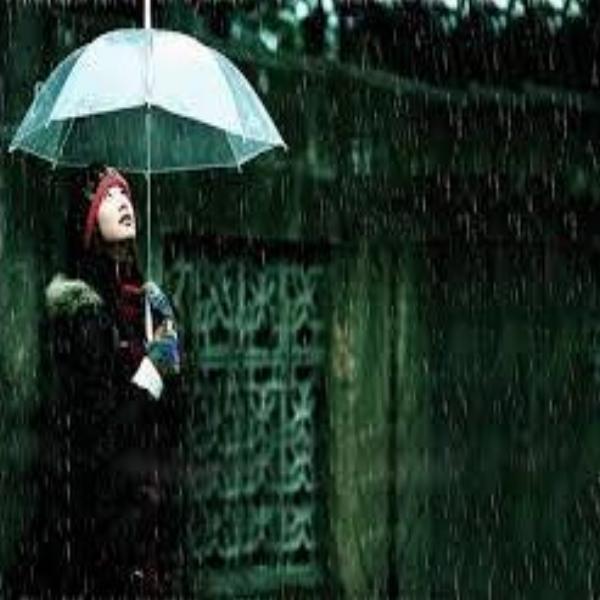}
	\includegraphics[width=1.06in]{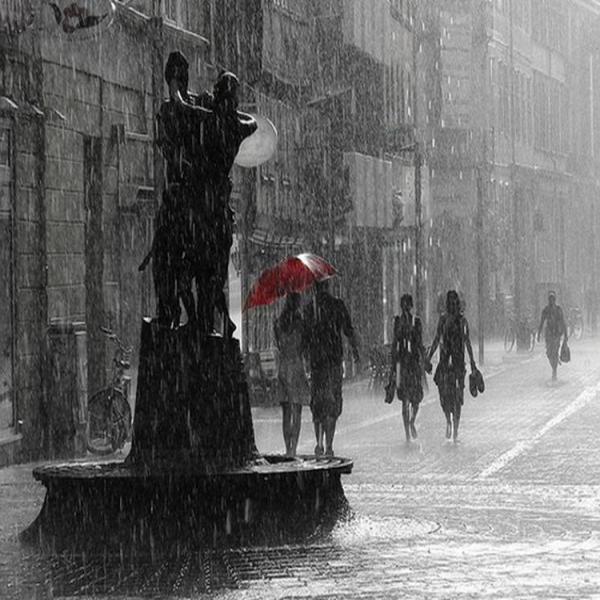}
	\includegraphics[width=1.06in]{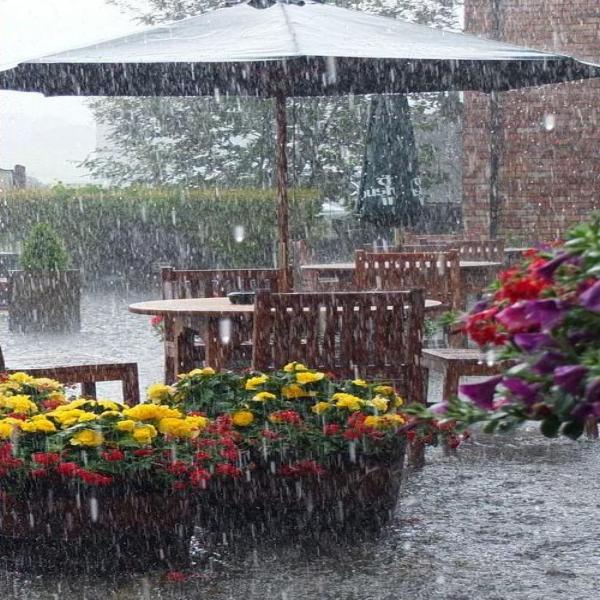}
	\includegraphics[width=1.06in]{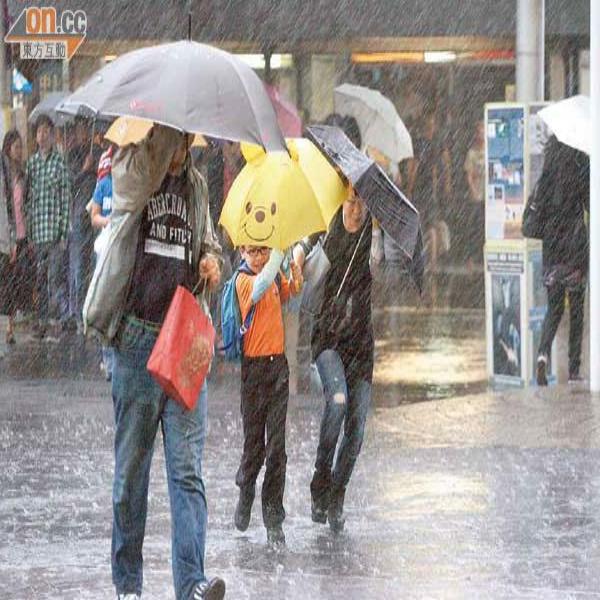}
	\caption{Sample images from real-world rainy dataset.}
	\label{fig:realdataset}
\end{figure*}

\subsection{Refined Perceptual Loss}
\begin{figure*}[!]
	\centering
	\begin{minipage}{.135\textwidth}
		\centering
		\includegraphics[width=2.5cm,height=1.7cm]{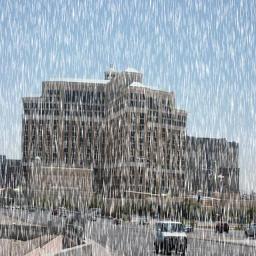}
		\captionsetup{labelformat=empty}
		\captionsetup{justification=centering}
		%\vskip+6pt
	\end{minipage}
	\begin{minipage}{.135\textwidth}
		\centering
		\includegraphics[width=2.5cm,height=1.7cm]{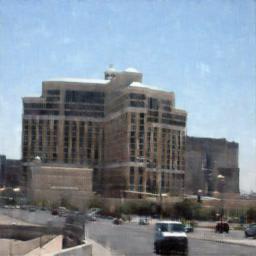}
		\captionsetup{labelformat=empty}
		\captionsetup{justification=centering}
		%\vskip+6pt
	\end{minipage}
	\begin{minipage}{.135\textwidth}
		\centering
		\includegraphics[width=2.5cm,height=1.7cm]{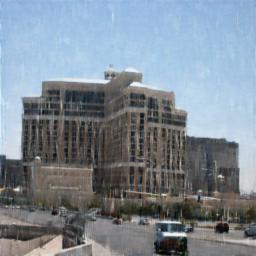}
		\captionsetup{labelformat=empty}
		\captionsetup{justification=centering}
		%\vskip+6pt
	\end{minipage}
	\begin{minipage}{.135\textwidth}
		\centering
		\includegraphics[width=2.5cm,height=1.7cm]{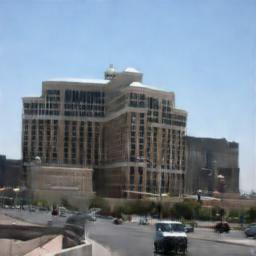}
		\captionsetup{labelformat=empty}
		\captionsetup{justification=centering}
		%\vskip+6pt
	\end{minipage}
	\begin{minipage}{.135\textwidth}
		\centering
		\includegraphics[width=2.5cm,height=1.7cm]{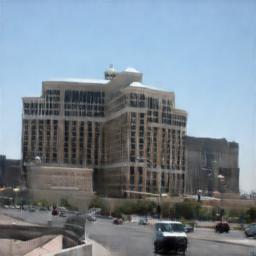}
		\captionsetup{labelformat=empty}
		\captionsetup{justification=centering}
		%\vskip+6pt
	\end{minipage}	 
	\begin{minipage}{.135\textwidth}
		\includegraphics[width=2.5cm,height=1.7cm]{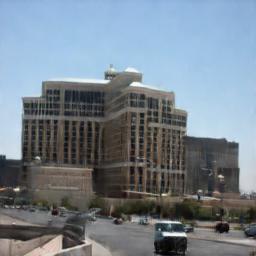}
		\captionsetup{labelformat=empty}
		\captionsetup{justification=centering}
		%\vskip+6pt
	\end{minipage}
	\begin{minipage}{.135\textwidth}
		\centering
		\includegraphics[width=2.5cm,height=1.7cm]{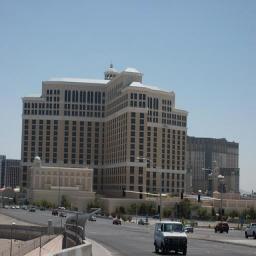}
		\captionsetup{labelformat=empty}
		\captionsetup{justification=centering}
		%\vskip+6pt
	\end{minipage}	\\		\vskip+6pt
	\begin{minipage}{.135\textwidth}
		\centering
		\includegraphics[width=2.5cm,height=1.2cm]{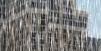}
		\captionsetup{labelformat=empty}
		\captionsetup{justification=centering}
		%\vskip+6pt
	\end{minipage}
	\begin{minipage}{.135\textwidth}
		\centering
		\includegraphics[width=2.5cm,height=1.2cm]{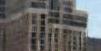}
		\captionsetup{labelformat=empty}
		\captionsetup{justification=centering}
		%\vskip+6pt
	\end{minipage}
	\begin{minipage}{.135\textwidth}
		\centering
		\includegraphics[width=2.5cm,height=1.2cm]{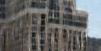}
		\captionsetup{labelformat=empty}
		\captionsetup{justification=centering}
		%\vskip+6pt
	\end{minipage}
	\begin{minipage}{.135\textwidth}
		\centering
		\includegraphics[width=2.5cm,height=1.2cm]{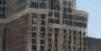}
		\captionsetup{labelformat=empty}
		\captionsetup{justification=centering}
		%\vskip+6pt
	\end{minipage}
	\begin{minipage}{.135\textwidth}
		\centering
		\includegraphics[width=2.5cm,height=1.2cm]{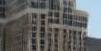}
		\captionsetup{labelformat=empty}
		\captionsetup{justification=centering}
		%\vskip+6pt
	\end{minipage}	 
	\begin{minipage}{.135\textwidth}
		\includegraphics[width=2.5cm,height=1.2cm]{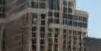}
		\captionsetup{labelformat=empty}
		\captionsetup{justification=centering}
		%\vskip+6pt
	\end{minipage}
	\begin{minipage}{.135\textwidth}
			\centering
			\includegraphics[width=2.5cm,height=1.2cm]{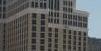}
			\captionsetup{labelformat=empty}
			\captionsetup{justification=centering}
			%\vskip+6pt
	\end{minipage}	\\		\vskip+6pt
	\begin{minipage}{.135\textwidth}
		\centering
		\includegraphics[width=2.5cm,height=1.2cm]{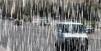}
		\captionsetup{labelformat=empty}
		\captionsetup{justification=centering}
		\caption*{(a) }
		%\vskip+6pt
	\end{minipage}
	\begin{minipage}{.135\textwidth}
		\centering
		\includegraphics[width=2.5cm,height=1.2cm]{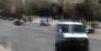}
		\captionsetup{labelformat=empty}
		\captionsetup{justification=centering}
		\caption*{(b)  }
		%\vskip+6pt
	\end{minipage}
	\begin{minipage}{.135\textwidth}
		\centering
		\includegraphics[width=2.5cm,height=1.2cm]{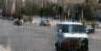}
		\captionsetup{labelformat=empty}
		\captionsetup{justification=centering}
		\caption*{(c) }
		%\vskip+6pt
	\end{minipage}
	\begin{minipage}{.135\textwidth}
		\centering
		\includegraphics[width=2.5cm,height=1.2cm]{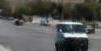}
		\captionsetup{labelformat=empty}
		\captionsetup{justification=centering}
		\caption*{(d) }
		%\vskip+6pt
	\end{minipage}
	\begin{minipage}{.135\textwidth}
		\centering
		\includegraphics[width=2.5cm,height=1.2cm]{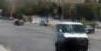}
		\captionsetup{labelformat=empty}
		\captionsetup{justification=centering}
		\caption*{(e) }
		%\vskip+6pt
	\end{minipage}	 
	\begin{minipage}{.135\textwidth}
		\includegraphics[width=2.5cm,height=1.2cm]{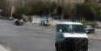}
		\captionsetup{labelformat=empty}
		\captionsetup{justification=centering}
		\caption*{(f) }
		%\vskip+6pt
	\end{minipage}
	\begin{minipage}{.135\textwidth}
		\centering
		\includegraphics[width=2.5cm,height=1.2cm]{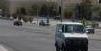}
		\captionsetup{labelformat=empty}
		\captionsetup{justification=centering}
		\caption*{(g)  }
		%\vskip+6pt
	\end{minipage}
	\caption{Qualitative comparisons for different baseline configurations of the proposed method. (a) Input image, (b) GEN, (c) GEN-CGAN-S, (d) GEN-P, (e) GEN-PS, (f) ID-CGAN and (g) Target image.  }
	\label{fig:baseline_real}
\end{figure*}

As discussed earlier, GANs  are known to be unstable to train and they may produce noisy or incomprehensible results via the guided generator. A probable reason is that the new input may not come from the same distribution of the training samples. As illustrated in Fig.~\ref{fig:baseline_real}(c), it can be clearly observed that there are some artifacts introduced by the normal GAN structure. This  greatly influences the visual performance of the output image. A possible solution to  address this issue is to introduce perceptual loss into the network. Recently, loss function measured on the difference of  high-level feature representation, such as loss measured on certain layers in CNN \cite{perceptual_loss},  has demonstrated much better visual performance than the per-pixel loss used in traditional CNNs. However, in many cases it fails to preserve color and texture information \cite{perceptual_loss}. Also, it does not achieve good quantitative performance simultaneously. To ensure that the results have good visual and quantitative scores along with good discriminatory performance, we propose a new refined loss function. Specifically, we combine pixel-to-pixel Euclidean loss, perceptual loss \cite{perceptual_loss} and adversarial loss together with appropriate weights to form our new refined loss function. The new loss function is then defined as follows:
 \begin{equation}
 \label{eq:loss_all}
 \begin{split}
 L_{RP}= L_{E}+ \lambda_aL_A+\lambda_pL_P,
 \end{split}
  \end{equation}
where $L_A$ represents adversarial loss (loss from the discriminator $D$), $L_P$ is perceptual loss and $L_{E}$ is normal per-pixel loss function such as Euclidean loss. Here, $\lambda_p$ and $\lambda_a$ are pre-defined weights for perceptual loss and adversarial  loss, respectively. If we set both  $\lambda_p$ and $\lambda_a$ to be 0, then the network reduces to a normal CNN configuration, which aims to minimize only the Euclidean loss between output image and ground truth. If $\lambda_p$ is set to 0, then the network reduces to a normal GAN.  If $\lambda_a$ set to 0, then the network reduces to the structure proposed in \cite{perceptual_loss}.

The three loss functions $L_P$, $L_E$ and $L_A$ are defined as follows. Given an image pair $\{\mathbf{x},\mathbf{y}_b\}$ with $C$ channels, width $W$ and height $H$  (i.e. $C\times W\times H$), where $\mathbf{x}$ is the input image and $\mathbf{y}_b$ is the corresponding ground truth, the per-pixel Euclidean loss is defined as: 
 \begin{equation}\label{eq:pie_loss}
 \begin{split}
 L_E= \frac{1}{CWH}\sum_{c=1}^{C}\sum_{x=1}^{W}\sum_{y=1}^{H} \|\phi_E({\mathbf{x}})^{c,w,h}-(\mathbf{y}_b)^{c,w,h}\|_2^2,
 \end{split}
 \end{equation}
 where $\phi_E$ is the learned network $G$ for generating the de-rained output.
Suppose the outputs of certain high-level layer are with size $C_i\times W_i\times H_i$. Similarly, the perceptual loss is defined as 
 \begin{equation}\label{eq:perc_loss}
 \begin{split}
 L_P= \frac{1}{C_iW_iH_i}\sum_{c=1}^{C_i}\sum_{w=1}^{W_i}\sum_{h=1}^{H_i} \|V(\phi_E({\mathbf{x}}))^{c,w,h}-V
 (\mathbf{y}_b)^{c,w,h}\|_2^2,
 \end{split}
 \end{equation}
 where $V$ represents a non-linear CNN transformation. Similar to the idea proposed in \cite{perceptual_loss},  we aim to minimize the distance between high-level features. In our method, we compute the feature loss at layer relu2$\_$2 in VGG-16 model \cite{vgg}.\footnote{https://github.com/ruimashita/caffe-train/blob/master/vgg.train$\_$val.prototxt}

Given a set of $N$ de-rained images generated from the generator $\{\phi_E({\mathbf{x}})\}_{i=1}^N$, the entropy loss from the discriminator  to guide the generator  is defined as:
 \begin{equation}\label{eq:adv_loss}
 \begin{split}
 L_A= -\frac{1}{N}\sum_{i=1}^{N}\log(D(\phi_E({\mathbf{x}}))).
 \end{split}
 \end{equation}

\section{Experiments and Results}
\label{sec:exp}

In this section, we present details of the experiments and quality measures used to evaluate the proposed ID-CGAN method. We also discuss the dataset and training details followed by comparison of the proposed method against a set of baseline methods and recent state-of-the-art approaches.

\subsection{Experimental Details}
\subsubsection{Synthetic dataset}  Due to the lack of availability of large size datasets for training and evaluation of single image de-raining, we synthesized a new set of training and testing samples in our experiments. The training set consists of a total of 700 images,  where 500 images are randomly chosen from the first 800 images in the UCID dataset \cite{ucid}  and 200 images are randomly chosen from the BSD-500's training set \cite{bsd_dataset}. The test set consists of a total of 100 images, where 50 images are randomly chosen from the last 500 images in the UCID dataset and 50 images are randomly chosen from the test-set of the BSD-500 dataset \cite{bsd_dataset}. After the train and test sets are created, we add rain-streaks to these images by following the guidelines mentioned in \cite{derain_tip17} using Photoshop\footnote{http://www.photoshopessentials.com/photo-effects/rain/}. It is ensured that rain pixels of different intensities and orientations are added to generate a diverse training and test set. Note that the images with rain form the set of observed images and the corresponding clean images form the set of ground truth images. All the  training and test samples are resized to 256$\times$256.

\subsubsection{Real-world rainy images dataset}  In order to demonstrate the effectiveness of the proposed method on real-world data, we created a dataset of 50 rainy images downloaded from the Internet.  While creating this dataset, we took all possible care to ensure that the images collected were diverse in terms of content as well as intensity and orientation of the rain pixels. A few sample images from this dataset are shown in Fig.~\ref{fig:realdataset}. This dataset is used for evaluation (test) purpose only.

\subsubsection{Quality measures}
The following measures are used to evaluate the performance of different methods:  Peak Signal to Noise Ratio (PSNR), Structural Similarity Index (SSIM) \cite{ssim}, Universal Quality Index (UQI) \cite{UIQ} and Visual Information Fidelity (VIF) \cite{VIF}. Similar to previous methods \cite{rain_2016_gmm},  all of these quantitative measures are calculated using the luminance channel. Since we do not have ground truth reference images for the real dataset, the performance of the proposed and other methods on the real dataset is evaluated visually.

\subsection{Model Details and Parameters}
The entire network is trained  on a Nvidia Titan-X GPU using the torch framework\cite{torch}. We used a batch size of 1 and number of training iterations of 100k. Adam algorithm \cite{adam_opt} with a learning rate of $2\times10^{-3}$ is used. During training, we set $\lambda_a=6.6\times10^{-3}$ and $\lambda_p=1$. All the parameters are set via cross-validation. A low value for $\lambda_a$ is used so as to ensure that the adversarial loss does not dominate the other losses. 

\begin{table}[t!]
	\centering
	\resizebox{0.45\textwidth}{!}{%
		\begin{tabular}{cccccc}
			\hline
			& GEN & GEN-CGAN-S & GEN-P & GEN-CGAN-PS & ID-CGAN \\ \hline\hline
			PSNR (dB) & \textbf{24.40} & 23.55 & 23.77 & 24.08 & {24.34} \\ \hline
			SSIM & 0.8275 & 0.8290 & 0.8305 & 0.8376  &\textbf{0.8430} \\ \hline
			UQI & 0.6506 & 0.6541 & 0.6557 & 0.6705 &\textbf{0.6741}  \\ \hline
			VIF & 0.3999 & 0.4003 & 0.4056 & 0.4133 &\textbf{0.4188} \\ \hline
		\end{tabular}}
		\caption{Quantitative comparison baseline configurations.}
		\label{tab:baselinetable}
	\end{table}

%\begin{table}[t!]
%	\centering
%	\resizebox{0.45\textwidth}{!}{%
%		\begin{tabular}{cccccc}
%			\hline
%			&  Single-scale &  Two-scale &  Three-scale & Four-scale \\ \hline\hline
%			PSNR (dB) & 24.08  & 24.11 & 24.20  & \textbf{24.34} \\ \hline
%			SSIM &  0.8376   &  0.8386  &  0.8428   &\textbf{0.8430} \\ \hline
%			UQI & 0.6705 & 0.6722  & 0.6723 & \textbf{0.6741}  \\ \hline
%			VIF & 0.4133 & 0.4155  & 0.4184  &\textbf{0.4188} \\ \hline
%	\end{tabular}}
%	\caption{Quantitative comparison of  baseline configurations with different scales in discriminator.  }
%	\label{tab:baselinetable}
%\end{table}
	
%\begin{table}[t!]
%	\centering
%	\resizebox{0.45\textwidth}{!}{%
%		\begin{tabular}{cccccc}
%			\hline
%			&DDN & ID-CGAN (Vanilla) & ID-CGAN \\ \hline\hline
%			PSNR (dB) & 23.28 & 22.73 & {24.34} \\ \hline
%			SSIM  & 0.8208 & 0.8113  &\textbf{0.8430} \\ \hline
%			UQI  & 0.6533 & 0.6449 &\textbf{0.6741}  \\ \hline
%			VIF  & 0.4116 & 0.4148 &\textbf{0.4188} \\ \hline
%	\end{tabular}}
%	\caption{Quantitative comparisons.}
%\end{table}	

\begin{table*}[ht!]
	\centering
	\resizebox{0.9\textwidth}{!}{%
		\begin{tabular}{ccccccccccc}
			\hline
			& SPM \cite{derain_tip12}  & PRM \cite{derain_lowrank} & DSC \cite{dis_rain_2015} & CNN \cite{derain_tip17} & GMM \cite{rain_2016_gmm} & CCR \cite{rain_wacv2017} & DDN \cite{derain_cvpr2017} & JORDER \cite{derain_cvpr2017_multi} & PAN \cite{wang2018perceptual} &ID-CGAN \\ \hline\hline
			PSNR (dB) & 18.88 & 20.46 & 18.56 & 19.12 & 22.27 & 20.56 & 23.28  & 21.09 & 23.35 &\textbf{24.34} \\ \hline
			SSIM & 0.7632 & 0.7297 & 0.5996 & 0.6013 & 0.7413 & 0.7332 & 0.8208 & 0.7525 & 0.8303 &\textbf{0.8430} \\ \hline
			UQI & 0.4149 & 0.5668 & 0.4804 & 0.4706 & 0.5751 & 0.5582 &  0.6533 & 0.5768 & 0.6644 & \textbf{0.6741} \\ \hline
			VIF & 0.2197 & 0.3441 & 0.3325 & 0.3307 & 0.4042 & 0.3607 & 0.4116 & 0.3785 & 0.4050 &\textbf{0.4188} \\ \hline
		\end{tabular}
	}
	\caption{Quantitative comparisons with state-of-the-art methods evaluated on using four different criterions.  }
	\label{ta:quantitive}
\end{table*}

\subsection{Comparison with Baseline Configurations}
In order to demonstrate the significance of different modules in the proposed method, we compare the performance of following baseline configurations:
\begin{itemize}[topsep=0pt,noitemsep,leftmargin=*]
  \item {GEN}: Generator $G$ is trained using per-pixel Euclidean loss by setting $\lambda_a$ and $\lambda_p$ to zero in \eqref{eq:loss_all}. This amounts to a traditional CNN architecture with Euclidean loss. 
  \item {GEN-CGAN-S}: Generator $G$ trained using per-pixel Euclidean loss and Adversarial loss from a single-scale discriminator $D$ (no multi-scale pooling). $\lambda_p$ is set to zero in \eqref{eq:loss_all}. 
  \item {GEN-P}: Generator $G$ is trained using per-pixel Euclidean loss and perceptual loss. $\lambda_a$ is set to zero in \eqref{eq:loss_all}.  
  \item {GEN-CGAN-PS}:  Generator $G$ is trained using per-pixel Euclidean loss, perceptual loss and adversarial loss from a single scale discriminator. 
  \item {ID-CGAN}: Generator $G$ is trained using per-pixel Euclidean loss, perceptual loss and adversarial loss from multi-scale discriminator $D$.    
\end{itemize}

All four configurations along with ID-CGAN are learned using training images from the synthetic  training dataset. Results of quantitative performance, using the measures discussed earlier on test images from the synthetic dataset, are shown in Table \ref{tab:baselinetable}. Sample results for the above baseline configurations on test images from real dataset are shown in Fig.~\ref{fig:baseline_real} and Fig.~\ref{fig:baseline_real1}.  It can be observed from Fig. \ref{fig:baseline_real}(c), that the introduction of adversarial loss improves the visual quality over the traditional CNN architectures, however, it also introduces certain artifacts.  The use of perceptual loss along with adversarial loss from a single scale discriminator reduces these artifacts while producing sharper  results. However, part of the texture details are still missing in the de-rained results (Fig.~\ref{fig:baseline_real}(e)) such as the edge of the left back part of the car (shown in third row in Fig.~\ref{fig:baseline_real}) and the structure of truck's front mirror (shown in second row in Fig.~\ref{fig:baseline_real1}).   Finally, the use of adversarial loss from multi-scale discriminator along with other loss functions (ID-CGAN) results in recovery of these texture details and achieve the best results.   Quantitative results shown in Table~\ref{ta:quantitive} also demonstrate the effectiveness of the each module.

\begin{figure*}[htp!]
	\centering
	\begin{minipage}{.18\textwidth}
		\centering
		\includegraphics[width=1\textwidth]{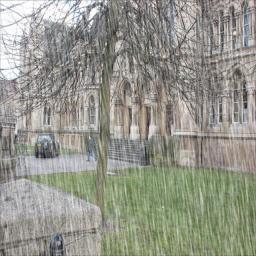}
		\captionsetup{labelformat=empty}
		\captionsetup{justification=centering}
		%\vskip+6pt
		\caption*{Input}
	\end{minipage}  
	\begin{minipage}{.18\textwidth}
		\centering
		\includegraphics[width=1\textwidth]{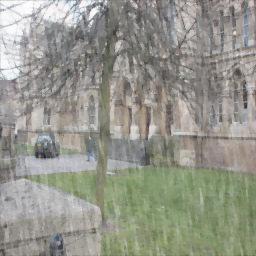}
		\captionsetup{labelformat=empty}
		\captionsetup{justification=centering}
		%\vskip+6pt
		\caption*{PRM \cite{derain_lowrank} }
	\end{minipage} 
	\begin{minipage}{.18\textwidth}
		\centering
		\includegraphics[width=1\textwidth]{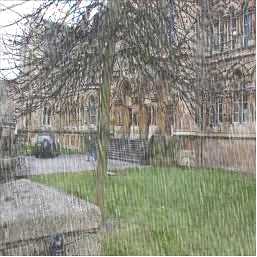}
		\captionsetup{labelformat=empty}
		\captionsetup{justification=centering}
		%\vskip+6pt
		\caption*{DSC \cite{dis_rain_2015}}
	\end{minipage}
	\begin{minipage}{.18\textwidth}
		\centering
		\includegraphics[width=1\textwidth]{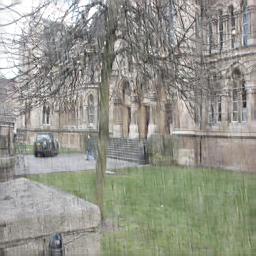}
		\captionsetup{labelformat=empty}
		\captionsetup{justification=centering}
		%\vskip+6pt
		\caption*{CNN \cite{derain_tip17} }
	\end{minipage}
	\begin{minipage}{.18\textwidth}
		\centering
		\includegraphics[width=1\textwidth]{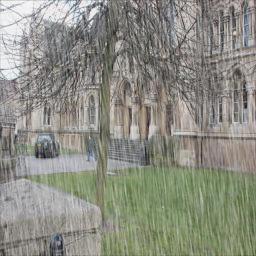}
		\captionsetup{labelformat=empty}
		\captionsetup{justification=centering}
		%\vskip+6pt
		\caption*{GMM \cite{rain_2016_gmm}}
	\end{minipage}
	\begin{minipage}{.18\textwidth}
		\centering
		\includegraphics[width=1\textwidth]{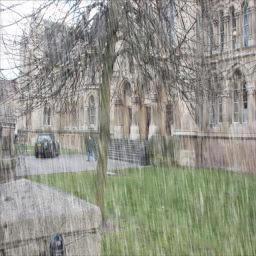}
		\captionsetup{labelformat=empty}
		\captionsetup{justification=centering}
		%\vskip+6pt
		\caption*{CCR \cite{rain_wacv2017}}
	\end{minipage}
	\begin{minipage}{.18\textwidth}
		\centering
		\includegraphics[width=1\textwidth]{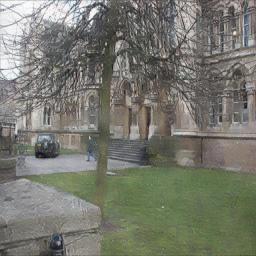}
		\captionsetup{labelformat=empty}
		\captionsetup{justification=centering}
		%\vskip+6pt
		\caption* {DDN \cite{derain_cvpr2017}}
	\end{minipage} 
	\begin{minipage}{.18\textwidth}
		\centering
		\includegraphics[width=1\textwidth]{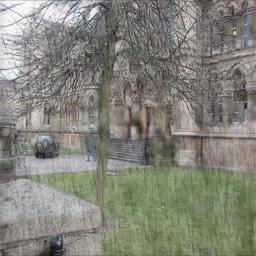}
		\captionsetup{labelformat=empty}
		\captionsetup{justification=centering}
		%\vskip+6pt
		\caption*{JORDER \cite{derain_cvpr2017_multi}}
	\end{minipage}
	\begin{minipage}{.18\textwidth}
		\centering
		\includegraphics[width=1\textwidth]{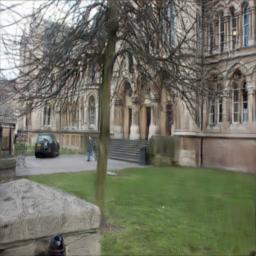}
		\captionsetup{labelformat=empty}
		\captionsetup{justification=centering}
		%\vskip+6pt
		\caption* {ID-CGAN}
	\end{minipage} 
	\begin{minipage}{.18\textwidth}
		\centering
		\includegraphics[width=1\textwidth]{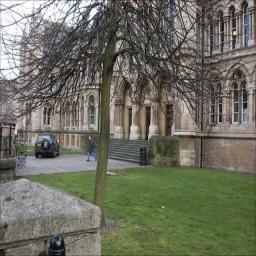}
		\captionsetup{labelformat=empty}
		\captionsetup{justification=centering}
		%\vskip+6pt
		\caption*{Ground Truth}
	\end{minipage} \\ \vskip+6pt
	\begin{minipage}{.18\textwidth}
		\centering
		\includegraphics[width=1\textwidth]{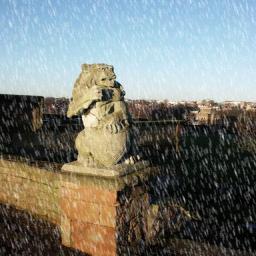}
		\captionsetup{labelformat=empty}
		\captionsetup{justification=centering}
		%\vskip+6pt
		\caption*{Input}
	\end{minipage} 	 
	\begin{minipage}{.18\textwidth}
		\centering
		\includegraphics[width=1\textwidth]{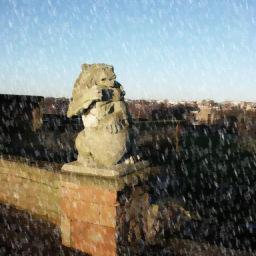}
		\captionsetup{labelformat=empty}
		\captionsetup{justification=centering}
		%\vskip+6pt
		\caption*{PRM \cite{derain_lowrank} }
	\end{minipage} 
	\begin{minipage}{.18\textwidth}
		\centering
		\includegraphics[width=1\textwidth]{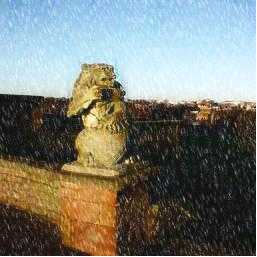}
		\captionsetup{labelformat=empty}
		\captionsetup{justification=centering}
		%\vskip+6pt
		\caption*{DSC \cite{dis_rain_2015}}
	\end{minipage}
	\begin{minipage}{.18\textwidth}
		\centering
		\includegraphics[width=1\textwidth]{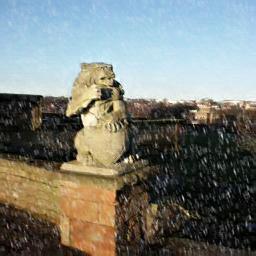}
		\captionsetup{labelformat=empty}
		\captionsetup{justification=centering}
		%\vskip+6pt
		\caption*{CNN \cite{derain_tip17} }
	\end{minipage}
	\begin{minipage}{.18\textwidth}
		\centering
		\includegraphics[width=1\textwidth]{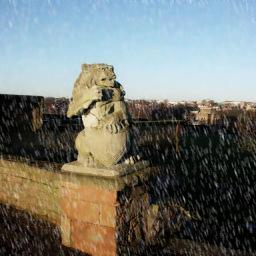}
		\captionsetup{labelformat=empty}
		\captionsetup{justification=centering}
		%\vskip+6pt
		\caption*{GMM \cite{rain_2016_gmm}}
	\end{minipage}
	\begin{minipage}{.18\textwidth}
		\centering
		\includegraphics[width=1\textwidth]{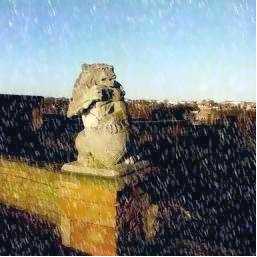}
		\captionsetup{labelformat=empty}
		\captionsetup{justification=centering}
		%\vskip+6pt
		\caption*{CCR \cite{rain_wacv2017}}
	\end{minipage}
	\begin{minipage}{.18\textwidth}
		\centering
		\includegraphics[width=1\textwidth]{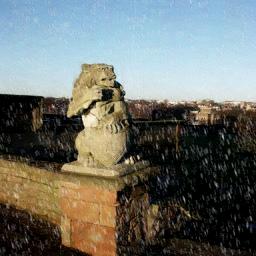}
		\captionsetup{labelformat=empty}
		\captionsetup{justification=centering}
		%\vskip+6pt
		\caption*{DDN \cite{derain_cvpr2017}}
	\end{minipage}
	\begin{minipage}{.18\textwidth}
		\centering
		\includegraphics[width=1\textwidth]{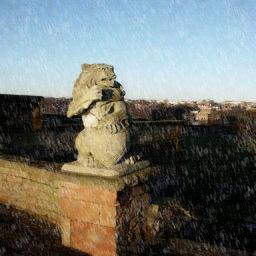}
		\captionsetup{labelformat=empty}
		\captionsetup{justification=centering}
		%\vskip+6pt
		\caption*{JORDER \cite{derain_cvpr2017_multi}}
	\end{minipage}
	\begin{minipage}{.18\textwidth}
		\centering
		\includegraphics[width=1\textwidth]{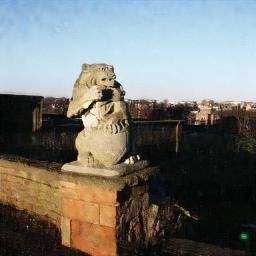}
		\captionsetup{labelformat=empty}
		\captionsetup{justification=centering}
		%\vskip+6pt
		\caption* {ID-CGAN}
	\end{minipage} 	
	\begin{minipage}{.18\textwidth}
		\centering
		\includegraphics[width=1\textwidth]{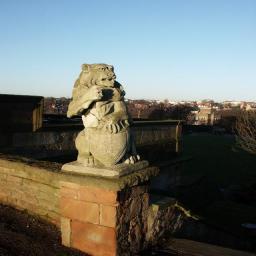}
		\captionsetup{labelformat=empty}
		\captionsetup{justification=centering}
		%\vskip+6pt
		\caption*{Ground Truth}
	\end{minipage} 
	
	\caption{Qualitative comparison of rain-streak removal on two sample images from synthetic dataset.}\label{syn_rainy}
\end{figure*}

\begin{figure*}[htp!]
	\centering
	\begin{minipage}{.24\textwidth}
		\centering
		\includegraphics[width=4.3cm,height=2.6cm]{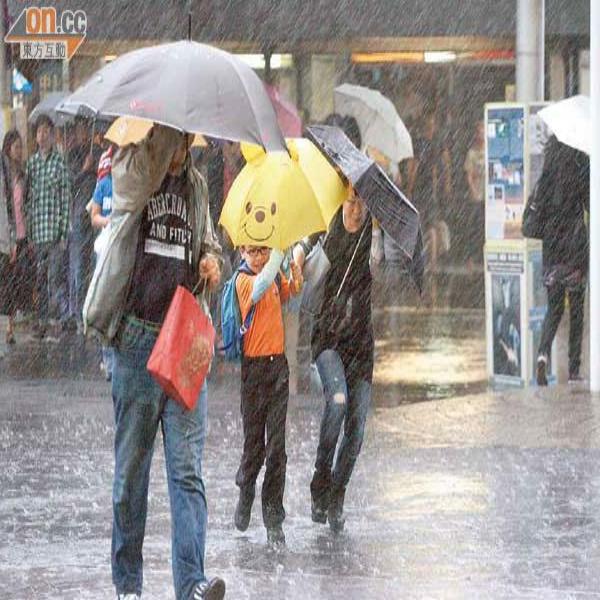}
		\captionsetup{labelformat=empty}
		\captionsetup{justification=centering}
		%\vskip+6pt
		\caption*{Input}
	\end{minipage} 
	\begin{minipage}{.24\textwidth}
		\centering
		\includegraphics[width=4.3cm,height=2.6cm]{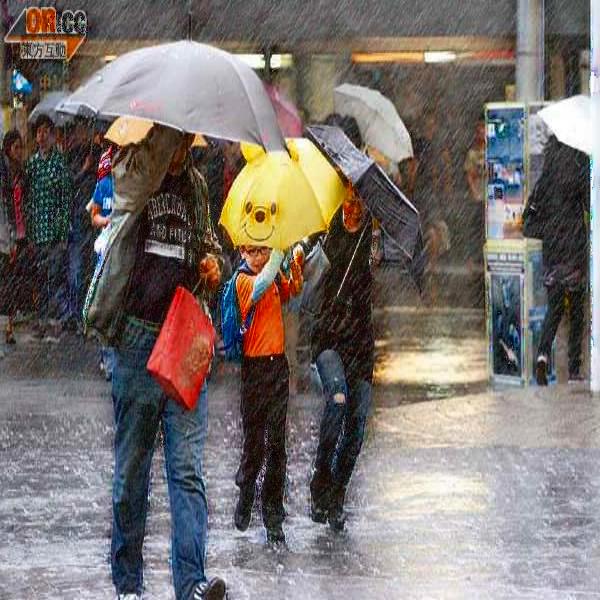}
		\captionsetup{labelformat=empty}
		\captionsetup{justification=centering}
		%\vskip+6pt
		\caption*{DSC \cite{dis_rain_2015}}
	\end{minipage}
	\begin{minipage}{.24\textwidth}
		\centering
		\includegraphics[width=4.3cm,height=2.6cm]{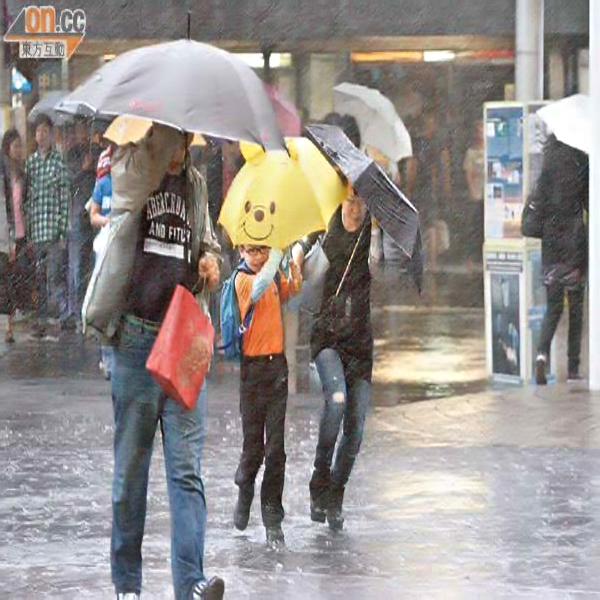}
		\captionsetup{labelformat=empty}
		\captionsetup{justification=centering}
		%\vskip+6pt
		\caption*{CNN \cite{derain_tip17}}
	\end{minipage} 
	\begin{minipage}{.24\textwidth}
		\centering
		\includegraphics[width=4.3cm,height=2.6cm]{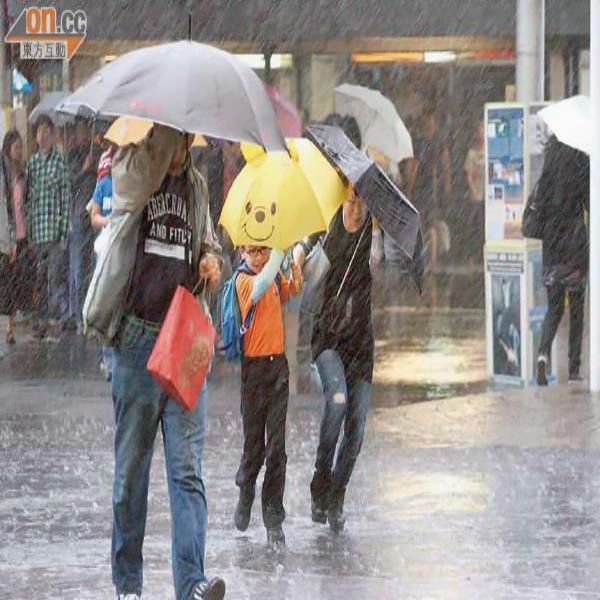}
		\captionsetup{labelformat=empty}
		\captionsetup{justification=centering}
		%\vskip+6pt
		\caption*{GMM \cite{rain_2016_gmm}}
	\end{minipage} \\
	\begin{minipage}{.24\textwidth}
		\centering
		\includegraphics[width=4.3cm,height=2.6cm]{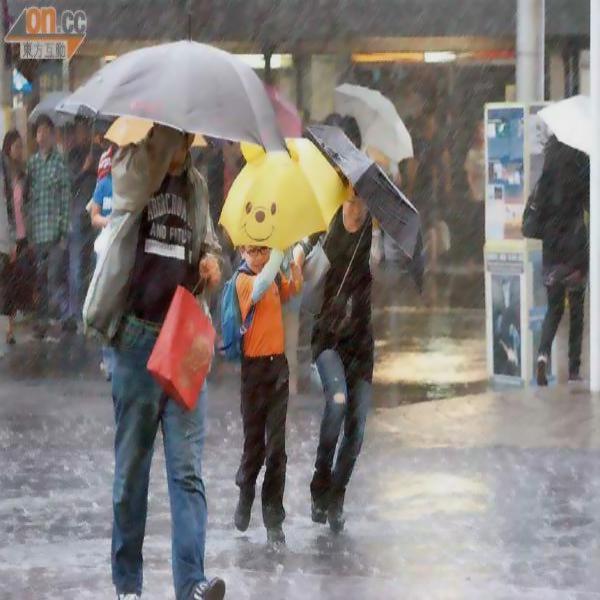}
		\captionsetup{labelformat=empty}
		\captionsetup{justification=centering}
		%\vskip+6pt
		\caption*{CCR \cite{rain_wacv2017} }
	\end{minipage}
	\begin{minipage}{.24\textwidth}
		\centering
		\includegraphics[width=4.3cm,height=2.6cm]{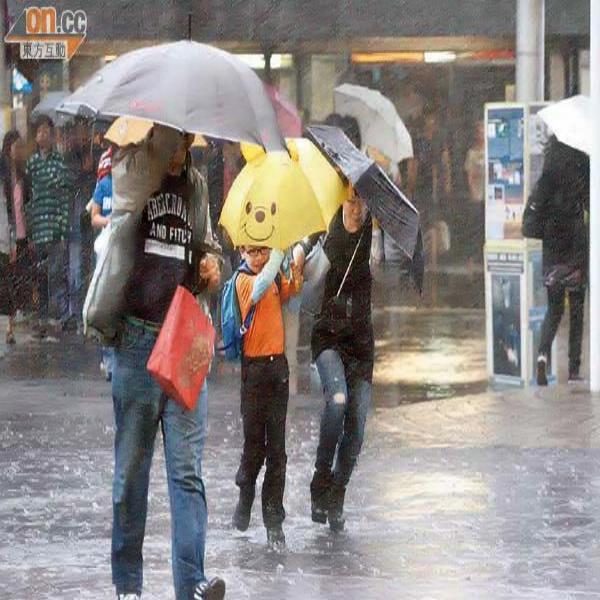}
		\captionsetup{labelformat=empty}
		\captionsetup{justification=centering}
		%\vskip+6pt
		\caption*{DDN \cite{derain_cvpr2017}}
	\end{minipage}
	\begin{minipage}{.24\textwidth}
		\centering
		\includegraphics[width=4.3cm,height=2.6cm]{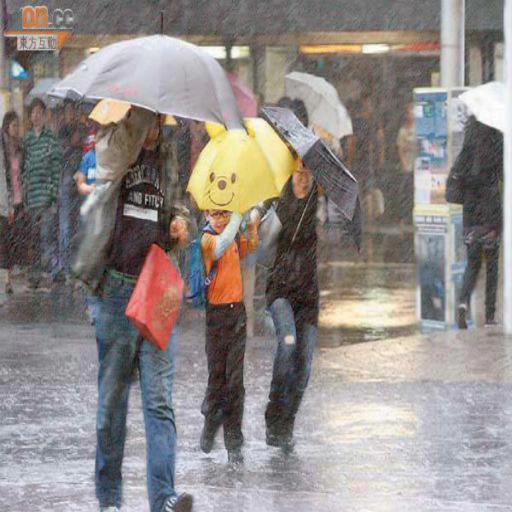}
		\captionsetup{labelformat=empty}
		\captionsetup{justification=centering}
		%\vskip+6pt
		\caption*{JORDER \cite{derain_cvpr2017_multi}}
	\end{minipage}
	\begin{minipage}{.24\textwidth}
		\centering
		\includegraphics[width=4.3cm,height=2.6cm]{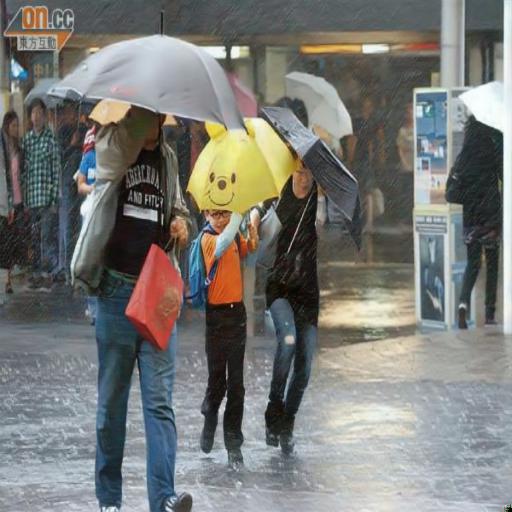}
		\captionsetup{labelformat=empty}
		\captionsetup{justification=centering}
		%\vskip+6pt
		\caption*{ID-CGAN}
	\end{minipage} 
	\\
	\begin{minipage}{.115\textwidth}
		\centering
		\includegraphics[width=2.15cm,height=1.3cm]{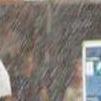}
		\captionsetup{labelformat=empty}
		\captionsetup{justification=centering}
		%\vskip+6pt
		\caption*{Input}
	\end{minipage}
	\begin{minipage}{.115\textwidth}
		\centering
		\includegraphics[width=2.15cm,height=1.3cm]{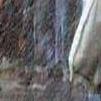}
		\captionsetup{labelformat=empty}
		\captionsetup{justification=centering}
		%\vskip+6pt
		\caption*{}
	\end{minipage}
	\begin{minipage}{.115\textwidth}
		\centering
		\includegraphics[width=2.15cm,height=1.3cm]{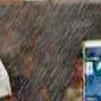}
		\captionsetup{labelformat=empty}
		\captionsetup{justification=centering}
		%\vskip+6pt
		\caption*{DSC \cite{dis_rain_2015}}
	\end{minipage}
	\begin{minipage}{.115\textwidth}
		\centering
		\includegraphics[width=2.15cm,height=1.3cm]{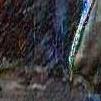}
		\captionsetup{labelformat=empty}
		\captionsetup{justification=centering}
		%\vskip+6pt
		\caption*{}
	\end{minipage}
	\begin{minipage}{.115\textwidth}
		\centering
		\includegraphics[width=2.15cm,height=1.3cm]{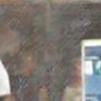}
		\captionsetup{labelformat=empty}
		\captionsetup{justification=centering}
		%\vskip+6pt
		\caption*{CNN \cite{derain_tip17}}
	\end{minipage}
	\begin{minipage}{.115\textwidth}
		\centering
		\includegraphics[width=2.15cm,height=1.3cm]{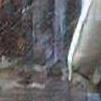}
		\captionsetup{labelformat=empty}
		\captionsetup{justification=centering}
		%\vskip+6pt
		\caption*{}
	\end{minipage}
	\begin{minipage}{.115\textwidth}
		\centering
		\includegraphics[width=2.15cm,height=1.3cm]{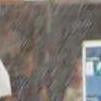}
		\captionsetup{labelformat=empty}
		\captionsetup{justification=centering}
		%\vskip+6pt
		\caption*{GMM \cite{rain_2016_gmm}}
	\end{minipage}
	\begin{minipage}{.115\textwidth}
		\centering
		\includegraphics[width=2.15cm,height=1.3cm]{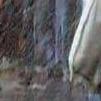}
		\captionsetup{labelformat=empty}
		\captionsetup{justification=centering}
		%\vskip+6pt
		\caption*{}
	\end{minipage}
	\\
\subfigure[]{
	\begin{minipage}{.115\textwidth}
		\centering
		\includegraphics[width=2.15cm,height=1.3cm]{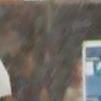}
		\captionsetup{labelformat=empty}
		\captionsetup{justification=centering}
		%\vskip+6pt
		\caption*{CCR \cite{rain_wacv2017} }
	\end{minipage}
	\begin{minipage}{.115\textwidth}
		\centering
		\includegraphics[width=2.15cm,height=1.3cm]{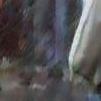}
		\captionsetup{labelformat=empty}
		\captionsetup{justification=centering}
		%\vskip+6pt
		\caption*{ }
	\end{minipage}
	\begin{minipage}{.115\textwidth}
		\centering
		\includegraphics[width=2.15cm,height=1.3cm]{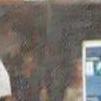}
		\captionsetup{labelformat=empty}
		\captionsetup{justification=centering}
		%\vskip+6pt
		\caption*{DDN \cite{derain_cvpr2017}}
	\end{minipage}
	\begin{minipage}{.115\textwidth}
		\centering
		\includegraphics[width=2.15cm,height=1.3cm]{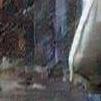}
		\captionsetup{labelformat=empty}
		\captionsetup{justification=centering}
		%\vskip+6pt
		\caption*{}
	\end{minipage}
	\begin{minipage}{.115\textwidth}
		\centering
		\includegraphics[width=2.15cm,height=1.3cm]{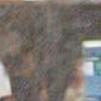}
		\captionsetup{labelformat=empty}
		\captionsetup{justification=centering}
		%\vskip+6pt
		\caption*{ JORDER \cite{derain_cvpr2017_multi}}
	\end{minipage}
	\begin{minipage}{.115\textwidth}
		\centering
		\includegraphics[width=2.15cm,height=1.3cm]{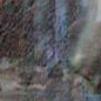}
		\captionsetup{labelformat=empty}
		\captionsetup{justification=centering}
		%\vskip+6pt
		\caption*{}
	\end{minipage}
	\begin{minipage}{.115\textwidth}
		\centering
		\includegraphics[width=2.15cm,height=1.3cm]{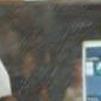}
		\captionsetup{labelformat=empty}
		\captionsetup{justification=centering}
		%\vskip+6pt
		\caption*{ID-CGAN}
	\end{minipage}
	\begin{minipage}{.115\textwidth}
		\centering
		\includegraphics[width=2.15cm,height=1.3cm]{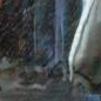}
		\captionsetup{labelformat=empty}
		\captionsetup{justification=centering}
		%\vskip+6pt
		\caption*{}
	\end{minipage}}
 \vskip+20pt

	\begin{minipage}{.24\textwidth}
		\centering
		\includegraphics[width=4.3cm,height=2.8cm]{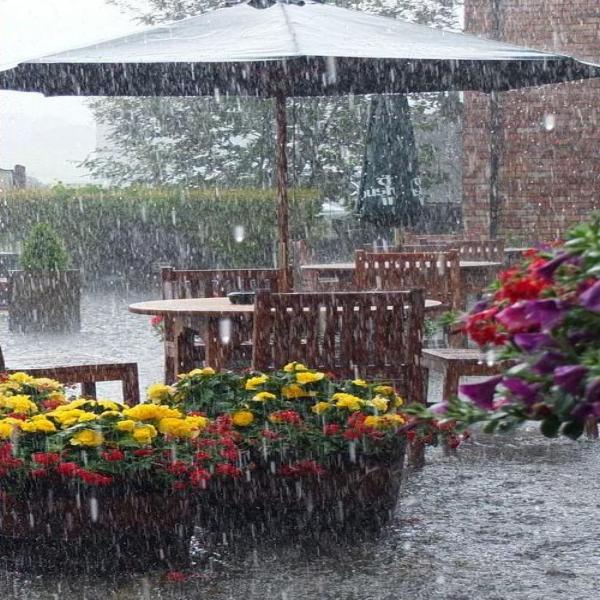}
		\captionsetup{labelformat=empty}
		\captionsetup{justification=centering}
		%\vskip+6pt
		\caption*{Input}
	\end{minipage} 	
	\begin{minipage}{.24\textwidth}
		\centering
		\includegraphics[width=4.3cm,height=2.8cm]{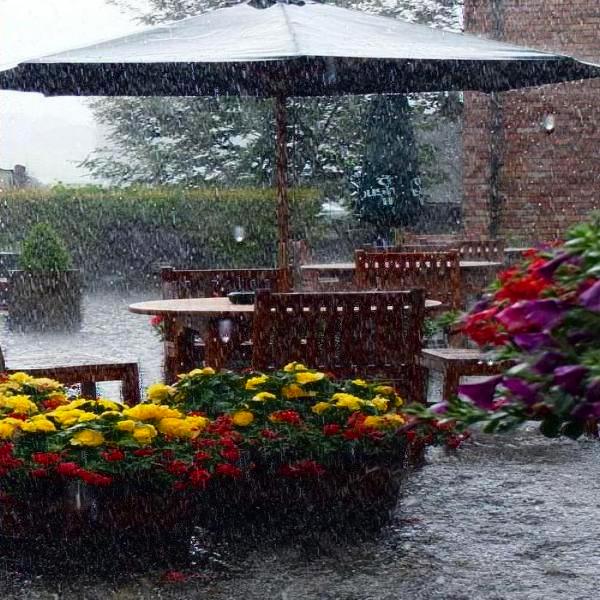}
		\captionsetup{labelformat=empty}
		\captionsetup{justification=centering}
		%\vskip+6pt
		\caption*{DSC \cite{dis_rain_2015}}
	\end{minipage} 	
		\begin{minipage}{.24\textwidth}
			\centering
			\includegraphics[width=4.3cm,height=2.8cm]{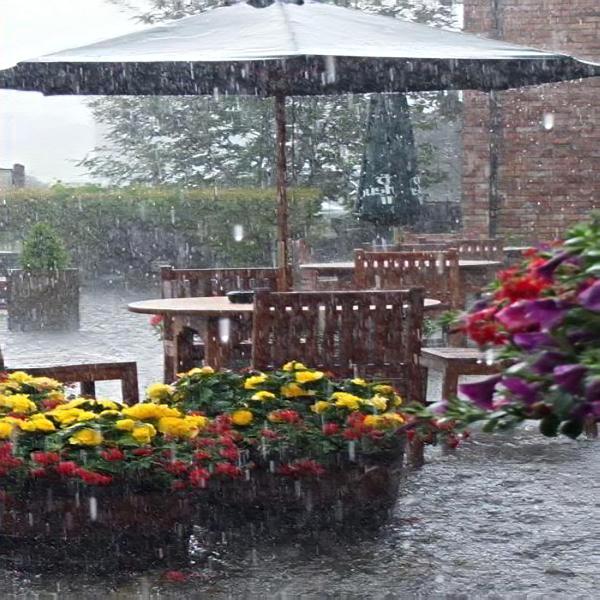}
			\captionsetup{labelformat=empty}
			\captionsetup{justification=centering}
			%\vskip+6pt
			\caption*{CNN \cite{derain_tip17}}
		\end{minipage} 	
			\begin{minipage}{.24\textwidth}
				\centering
				\includegraphics[width=4.3cm,height=2.8cm]{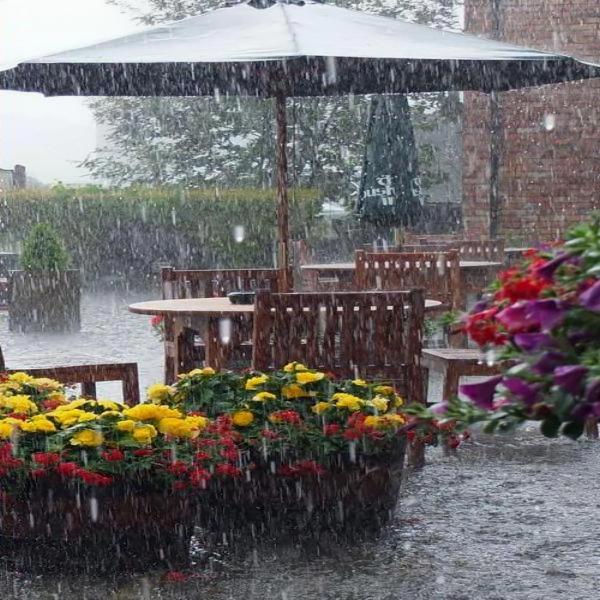}
				\captionsetup{labelformat=empty}
				\captionsetup{justification=centering}
				%\vskip+6pt
				\caption*{GMM \cite{rain_2016_gmm}}
			\end{minipage} 	\\
	\begin{minipage}{.24\textwidth}
		\centering
		\includegraphics[width=4.3cm,height=2.8cm]{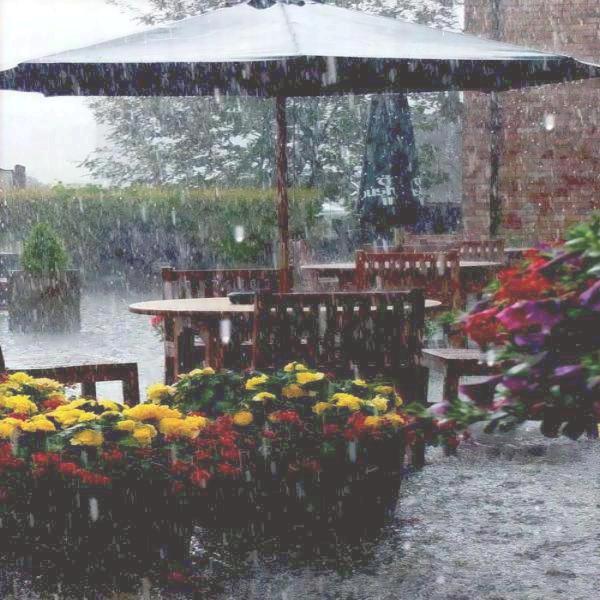}
		\captionsetup{labelformat=empty}
		\captionsetup{justification=centering}
		%\vskip+6pt
		\caption*{CCR \cite{rain_wacv2017} }
	\end{minipage} 	
		\begin{minipage}{.24\textwidth}
			\centering
			\includegraphics[width=4.3cm,height=2.8cm]{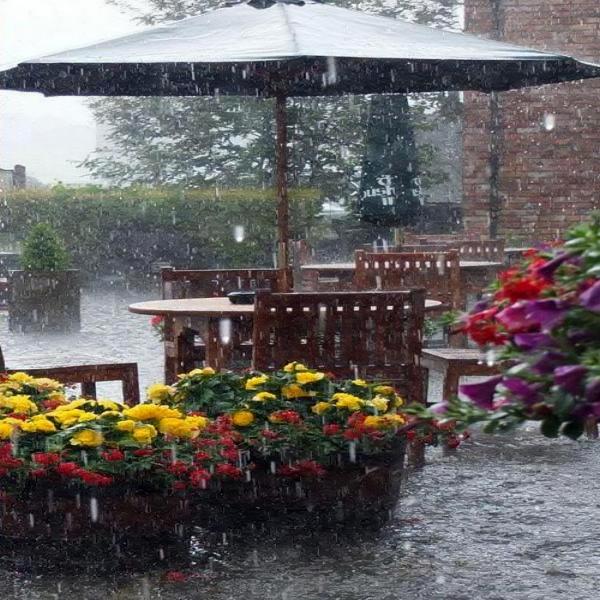}
			\captionsetup{labelformat=empty}
			\captionsetup{justification=centering}
			%\vskip+6pt
			\caption*{DDN \cite{derain_cvpr2017}}
		\end{minipage} 	
		\begin{minipage}{.24\textwidth}
			\centering
			\includegraphics[width=4.3cm,height=2.8cm]{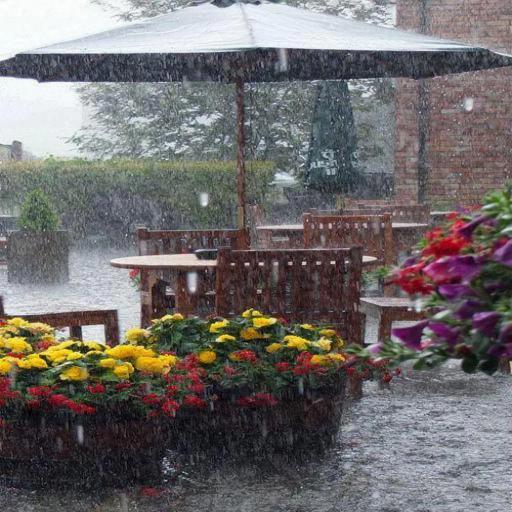}
			\captionsetup{labelformat=empty}
			\captionsetup{justification=centering}
			%\vskip+6pt
			\caption*{JORODR \cite{derain_cvpr2017_multi}}
		\end{minipage} 	
	\begin{minipage}{.24\textwidth}
		\centering
		\includegraphics[width=4.3cm,height=2.8cm]{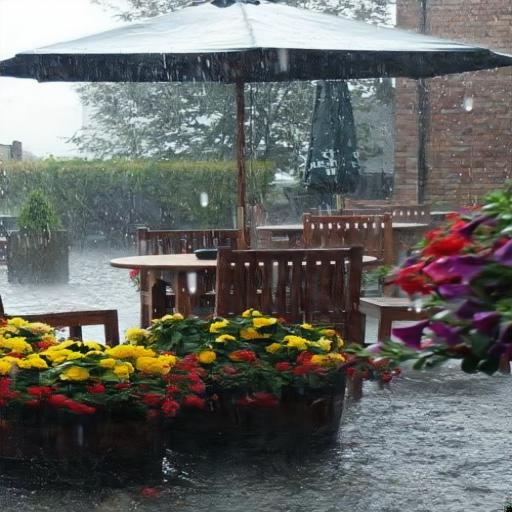}
		\captionsetup{labelformat=empty}
		\captionsetup{justification=centering}
		%\vskip+6pt
		\caption*{ID-CGAN}
	\end{minipage}\\
	\begin{minipage}{.115\textwidth}
		\centering
		\includegraphics[width=2.11cm,height=1.25cm]{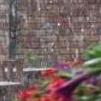}
		\captionsetup{labelformat=empty}
		\captionsetup{justification=centering}
		%\vskip+6pt
		\caption*{Input}
	\end{minipage}
	\begin{minipage}{.115\textwidth}
		\centering
		\includegraphics[width=2.11cm,height=1.25cm]{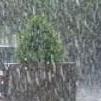}
		\captionsetup{labelformat=empty}
		\captionsetup{justification=centering}
		%\vskip+6pt
		\caption*{}
	\end{minipage}
	\begin{minipage}{.115\textwidth}
		\centering
		\includegraphics[width=2.11cm,height=1.25cm]{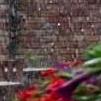}
		\captionsetup{labelformat=empty}
		\captionsetup{justification=centering}
		%\vskip+6pt
		\caption*{DSC \cite{dis_rain_2015}}
	\end{minipage}
	\begin{minipage}{.115\textwidth}
		\centering
		\includegraphics[width=2.11cm,height=1.25cm]{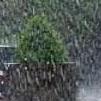}
		\captionsetup{labelformat=empty}
		\captionsetup{justification=centering}
		%\vskip+6pt
		\caption*{}
	\end{minipage}
	\begin{minipage}{.115\textwidth}
		\centering
		\includegraphics[width=2.11cm,height=1.25cm]{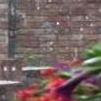}
		\captionsetup{labelformat=empty}
		\captionsetup{justification=centering}
		%\vskip+6pt
		\caption*{CNN \cite{derain_tip17}}
	\end{minipage}
	\begin{minipage}{.115\textwidth}
		\centering
		\includegraphics[width=2.11cm,height=1.25cm]{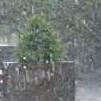}
		\captionsetup{labelformat=empty}
		\captionsetup{justification=centering}
		%\vskip+6pt
		\caption*{}
	\end{minipage}
	\begin{minipage}{.115\textwidth}
		\centering
		\includegraphics[width=2.11cm,height=1.25cm]{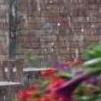}
		\captionsetup{labelformat=empty}
		\captionsetup{justification=centering}
		%\vskip+6pt
		\caption*{GMM \cite{rain_2016_gmm}}
	\end{minipage}
	\begin{minipage}{.115\textwidth}
		\centering
		\includegraphics[width=2.11cm,height=1.25cm]{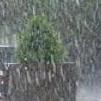}
		\captionsetup{labelformat=empty}
		\captionsetup{justification=centering}
		%\vskip+6pt
		\caption*{}
	\end{minipage}
	\\
\subfigure[]{
	\begin{minipage}{.115\textwidth}
		\centering
		\includegraphics[width=2.11cm,height=1.25cm]{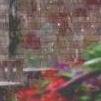}
		\captionsetup{labelformat=empty}
		\captionsetup{justification=centering}
		%\vskip+6pt
		\caption*{CCR \cite{rain_wacv2017}}
	\end{minipage}
	\begin{minipage}{.115\textwidth}
		\centering
		\includegraphics[width=2.11cm,height=1.25cm]{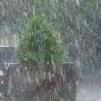}
		\captionsetup{labelformat=empty}
		\captionsetup{justification=centering}
		%\vskip+6pt
		\caption*{}
	\end{minipage}	
	\begin{minipage}{.115\textwidth}
		\centering
		\includegraphics[width=2.11cm,height=1.25cm]{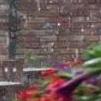}
		\captionsetup{labelformat=empty}
		\captionsetup{justification=centering}
		%\vskip+6pt
		\caption*{DDN \cite{derain_cvpr2017}}
	\end{minipage}
	\begin{minipage}{.115\textwidth}
		\centering
		\includegraphics[width=2.11cm,height=1.25cm]{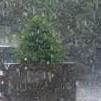}
		\captionsetup{labelformat=empty}
		\captionsetup{justification=centering}
		%\vskip+6pt
		\caption*{}
	\end{minipage}
	\begin{minipage}{.115\textwidth}
		\centering
		\includegraphics[width=2.11cm,height=1.25cm]{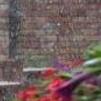}
		\captionsetup{labelformat=empty}
		\captionsetup{justification=centering}
		%\vskip+6pt
		\caption*{JORDER \cite{derain_cvpr2017_multi}}
	\end{minipage}
	\begin{minipage}{.115\textwidth}
		\centering
		\includegraphics[width=2.11cm,height=1.25cm]{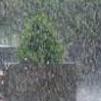}
		\captionsetup{labelformat=empty}
		\captionsetup{justification=centering}
		%\vskip+6pt
		\caption*{ }
	\end{minipage}
	\begin{minipage}{.115\textwidth}
		\centering
		\includegraphics[width=2.11cm,height=1.25cm]{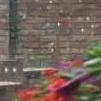}
		\captionsetup{labelformat=empty}
		\captionsetup{justification=centering}
		%\vskip+6pt
		\caption*{ID-CGAN}
	\end{minipage}
	\begin{minipage}{.115\textwidth}
		\centering
		\includegraphics[width=2.11cm,height=1.25cm]{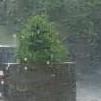}
		\captionsetup{labelformat=empty}
		\captionsetup{justification=centering}
		%\vskip+6pt
		\caption*{ }
	\end{minipage}}
 	
	\caption{Qualitative comparison of rain-streak removal on two sample real images.}\label{fig:color_ALL}
\end{figure*}

\subsection{Comparison with State-of-the-art Methods}
We compare the performance of the proposed ID-CGAN method with the following recent state-of-the-art methods for single image de-raining:
\begin{itemize}[topsep=0pt,noitemsep,leftmargin=*]
  \item {SPM}: Sparse dictionary-based method \cite{derain_tip12}  (\textit{TIP '12})
  \item {DSC}: Discriminative sparse coding-based method \cite{dis_rain_2015} (\textit{ICCV '15})
  \item {PRM}: PRM prior-based method\cite{rain_2016_gmm} (\textit{CVPR '16})
  \item {GMM}: GMM-based method \cite{derain_lowrank}  (\textit{ICCV '13})
  \item {CNN}: CNN-based method \cite{derain_tip17} (\textit{TIP '17})
  \item {CCR}: Convolutional-coding based method  \cite{rain_wacv2017} (\textit{WACV '17})
  \item {DDN}: Deep Detail Network method \cite{derain_cvpr2017} (\textit{CVPR '17})
  \item {JORDER}: CNN-based method  \cite{derain_cvpr2017_multi} (\textit{CVPR '17})
   \item {PAN}: GAN-based method  \cite{wang2018perceptual} (\textit{TIP '18})
\end{itemize}

\subsubsection{Evaluation on synthetic dataset}
In the first set of experiments, we evaluate the proposed method and compare its quantitative and qualitative performance against several state-of-the-art approaches  on  test images from the synthetic dataset. As the ground truth is available for the these test images, we calculate the quantitative measures such as PSNR, SSIM, UQI and VIF.   Table~\ref{ta:quantitive} shows the comparison of results based on these metrics. This table clear demonstrates  that the proposed ID-CGAN method is able to achieve superior quantitative performance as compared to the recent methods in terms of all the metrics stated earlier.

Fig.~\ref{syn_rainy} illustrates the qualitative improvements on two sample images from the synthetic dataset, achieved due to the use of the proposed method. Note that we selectively sample difficult images to show that our method performs well in difficult conditions. While PRM \cite{derain_lowrank} is able to remove the rain-streaks, it produces blurred results which are not visually appealing. The other compared methods are able to either reduce the intensity of rain or remove the streaks in parts, however, they fail to completely remove the rain-streaks. In contrast to the other methods, the proposed method is able to successfully remove majority of the rain streaks while maintaining the details of the de-rained images. 

\subsubsection{Evaluation on Real Rainy Images}
\begin{figure*}[htp!]
	\centering
	\begin{minipage}{.24\textwidth}
		\centering
		\includegraphics[width=1\textwidth]{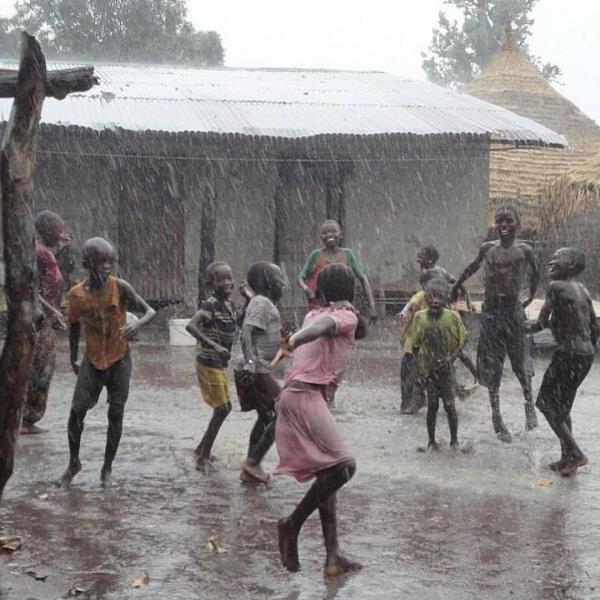}
		\captionsetup{labelformat=empty}
		\captionsetup{justification=centering}
		%\vskip+6pt
		\caption*{Input}
	\end{minipage}  
	\begin{minipage}{.24\textwidth}
		\centering
		\includegraphics[width=1\textwidth]{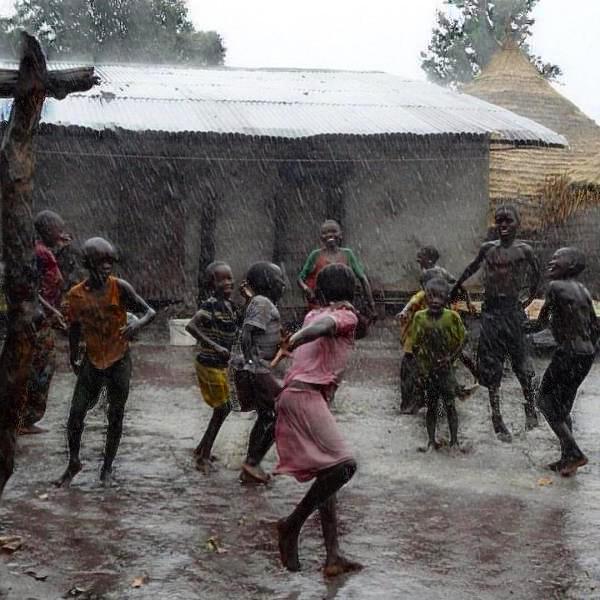}
		\captionsetup{labelformat=empty}
		\captionsetup{justification=centering}
		%\vskip+6pt
		\caption*{DSC \cite{dis_rain_2015}}
	\end{minipage}
	\begin{minipage}{.24\textwidth}
		\centering
		\includegraphics[width=1\textwidth]{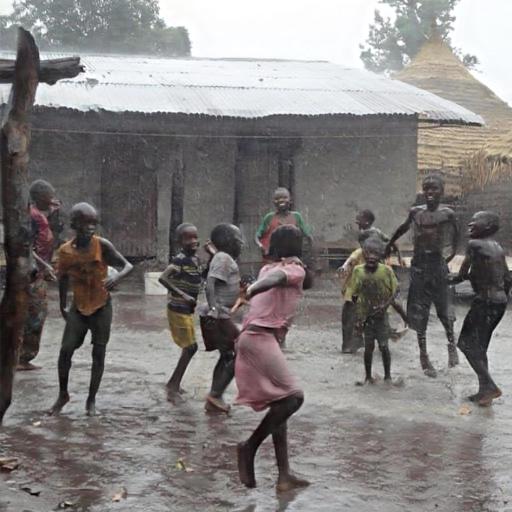}
		\captionsetup{labelformat=empty}
		\captionsetup{justification=centering}
		%\vskip+6pt
		\caption*{CNN \cite{derain_tip17} }
	\end{minipage}
	\begin{minipage}{.24\textwidth}
		\centering
		\includegraphics[width=1\textwidth]{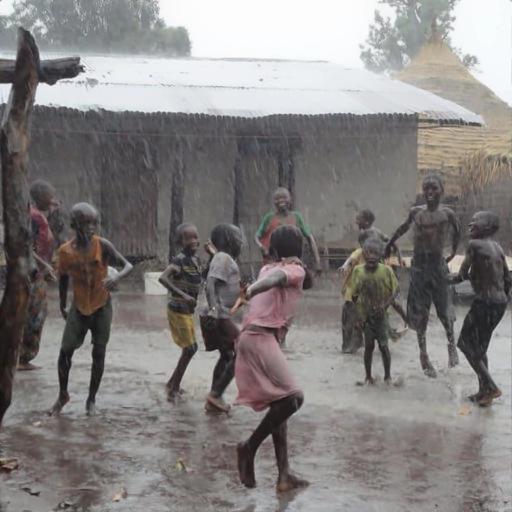}
		\captionsetup{labelformat=empty}
		\captionsetup{justification=centering}
		%\vskip+6pt
		\caption*{GMM \cite{rain_2016_gmm}}
	\end{minipage}
	\begin{minipage}{.24\textwidth}
		\centering
		\includegraphics[width=1\textwidth]{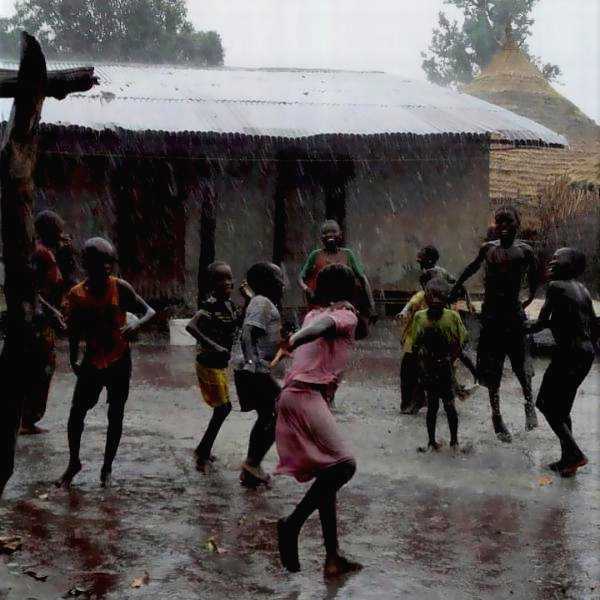}
		\captionsetup{labelformat=empty}
		\captionsetup{justification=centering}
		%\vskip+6pt
		\caption*{CCR \cite{rain_wacv2017}}
	\end{minipage}
	\begin{minipage}{.24\textwidth}
		\centering
		\includegraphics[width=1\textwidth]{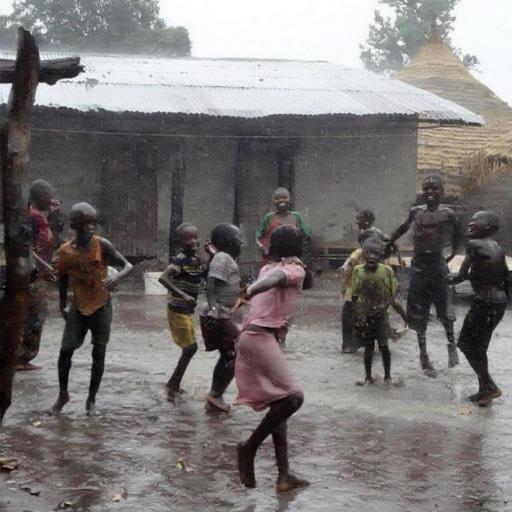}
		\captionsetup{labelformat=empty}
		\captionsetup{justification=centering}
		%\vskip+6pt
		\caption* {DDN \cite{derain_cvpr2017}}
	\end{minipage} 
	\begin{minipage}{.24\textwidth}
		\centering
		\includegraphics[width=1\textwidth]{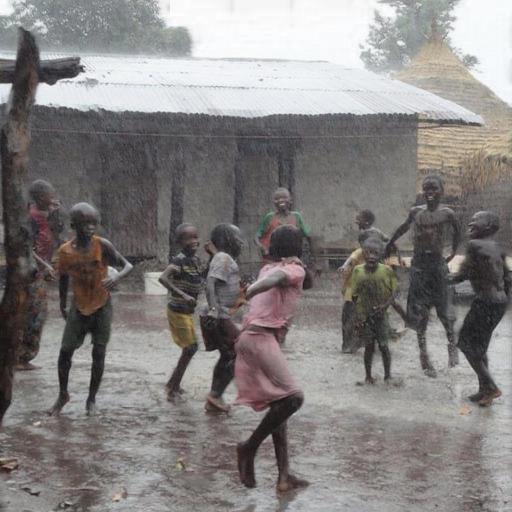}
		\captionsetup{labelformat=empty}
		\captionsetup{justification=centering}
		%\vskip+6pt
		\caption*{JORDER \cite{derain_tip12}}
	\end{minipage}
	\begin{minipage}{.24\textwidth}
		\centering
		\includegraphics[width=1\textwidth]{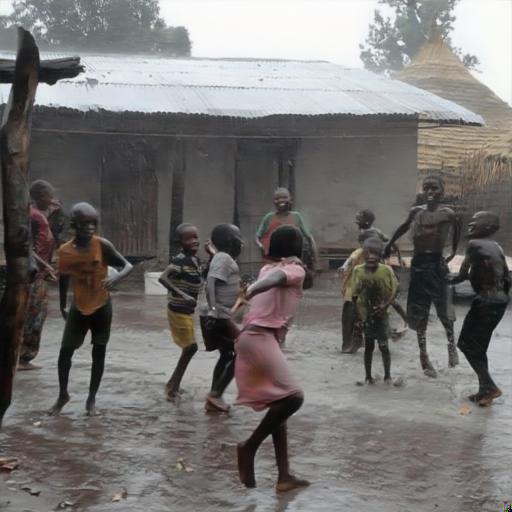}
		\captionsetup{labelformat=empty}
		\captionsetup{justification=centering}
		%\vskip+6pt
		\caption* {ID-CGAN}
	\end{minipage}  \\ \vskip+6pt
	\begin{minipage}{.24\textwidth}
		\centering
		\includegraphics[width=4.3cm, height=2.8cm]{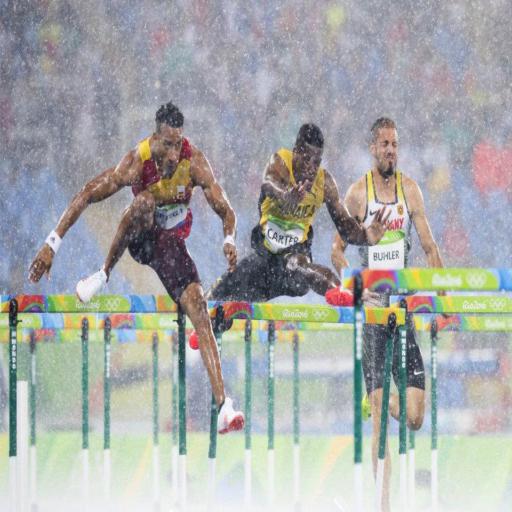}
		\captionsetup{labelformat=empty}
		\captionsetup{justification=centering}
		%\vskip+6pt
		\caption*{Input}
	\end{minipage} 	 
	\begin{minipage}{.24\textwidth}
		\centering
		\includegraphics[width=4.3cm, height=2.8cm]{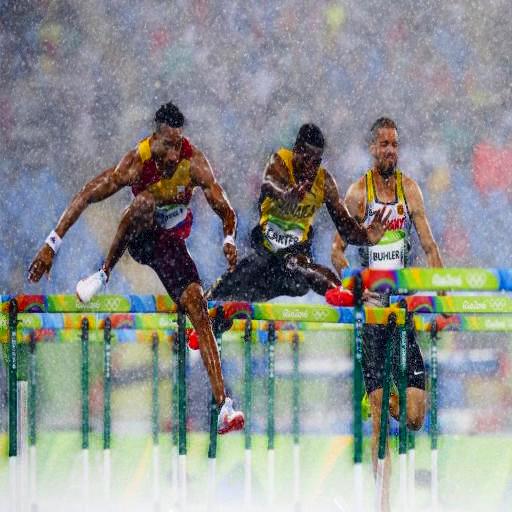}
		\captionsetup{labelformat=empty}
		\captionsetup{justification=centering}
		%\vskip+6pt
		\caption*{DSC \cite{dis_rain_2015}}
	\end{minipage}
	\begin{minipage}{.24\textwidth}
		\centering
			\includegraphics[width=4.3cm, height=2.8cm]{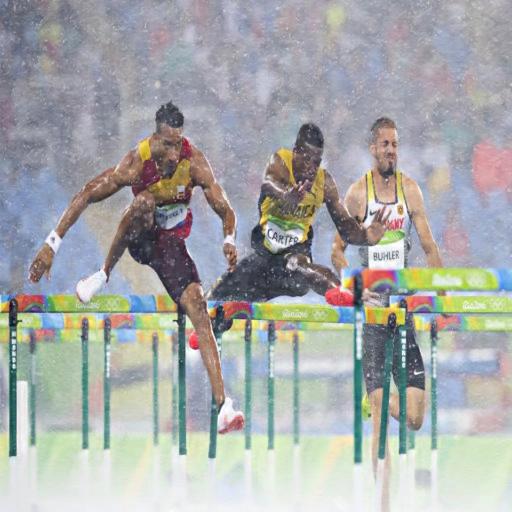}
		\captionsetup{labelformat=empty}
		\captionsetup{justification=centering}
		%\vskip+6pt
		\caption*{CNN \cite{derain_tip17} }
	\end{minipage}
	\begin{minipage}{.24\textwidth}
		\centering
		\includegraphics[width=4.3cm, height=2.8cm]{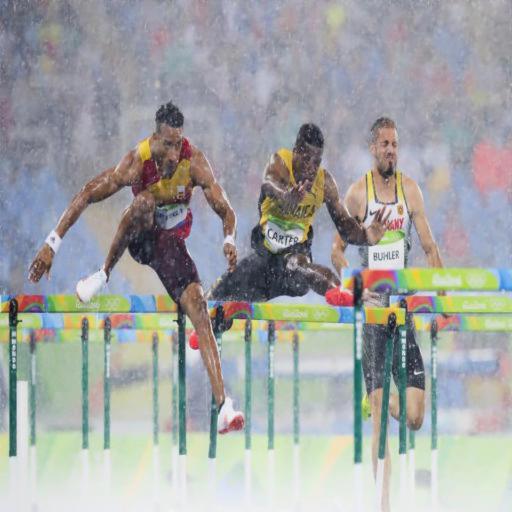}
		\captionsetup{labelformat=empty}
		\captionsetup{justification=centering}
		%\vskip+6pt
		\caption*{GMM \cite{rain_2016_gmm}}
	\end{minipage}
	\begin{minipage}{.24\textwidth}
		\centering
		\includegraphics[width=4.3cm, height=2.8cm]{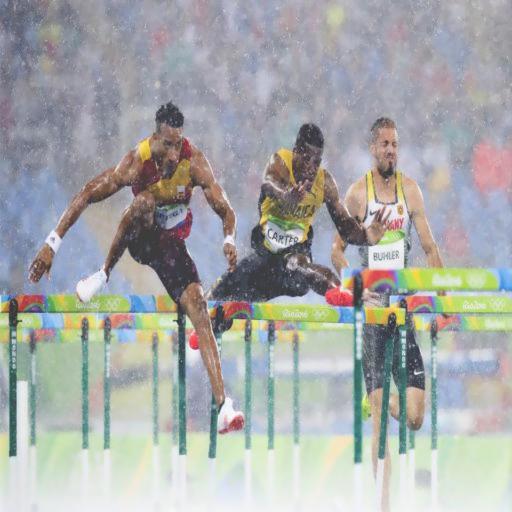}
		\captionsetup{labelformat=empty}
		\captionsetup{justification=centering}
		%\vskip+6pt
		\caption*{CCR \cite{rain_wacv2017}}
	\end{minipage}
	\begin{minipage}{.24\textwidth}
		\centering
		\includegraphics[width=4.3cm, height=2.8cm]{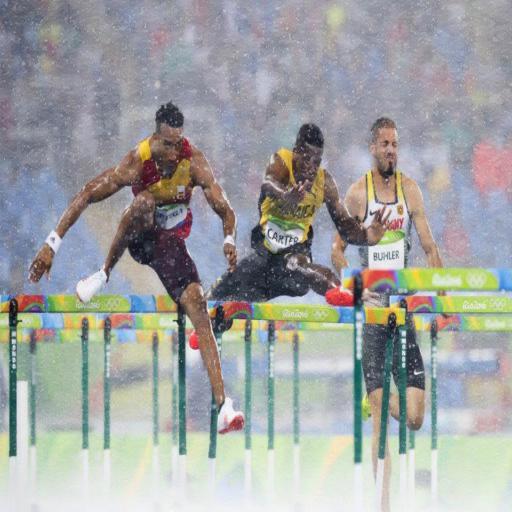}
		\captionsetup{labelformat=empty}
		\captionsetup{justification=centering}
		%\vskip+6pt
		\caption*{DDN \cite{derain_cvpr2017}}
	\end{minipage}
	\begin{minipage}{.24\textwidth}
		\centering
		\includegraphics[width=4.3cm, height=2.8cm]{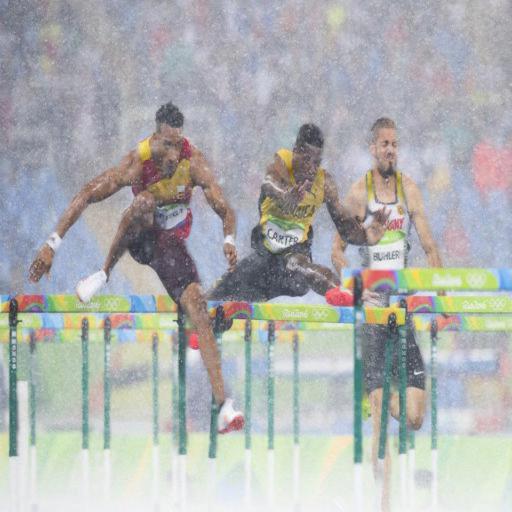}
		\captionsetup{labelformat=empty}
		\captionsetup{justification=centering}
		%\vskip+6pt
		\caption*{JORDER \cite{derain_cvpr2017_multi}}
	\end{minipage}
	\begin{minipage}{.24\textwidth}
		\centering
		\includegraphics[width=4.3cm, height=2.8cm]{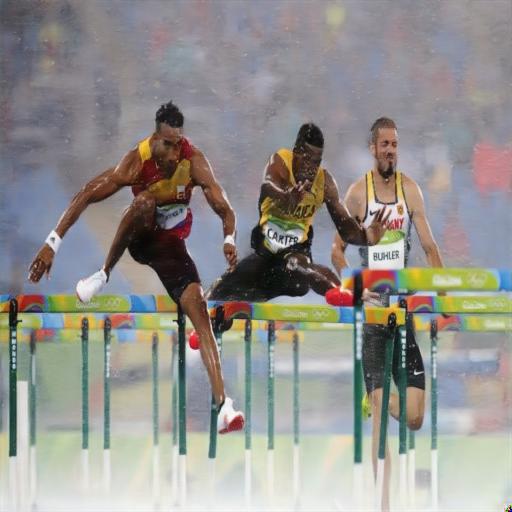}
		\captionsetup{labelformat=empty}
		\captionsetup{justification=centering}
		%\vskip+6pt
		\caption* {ID-CGAN}
	\end{minipage} 	
	
	\caption{Qualitative comparison of rain-streak removal on two sample real images.}\label{fig:real2}
\end{figure*}

\begin{figure}[t]
	\centering
	\begin{minipage}{.48\textwidth}
		\centering
		\includegraphics[width=8.3cm,height=2.6cm]{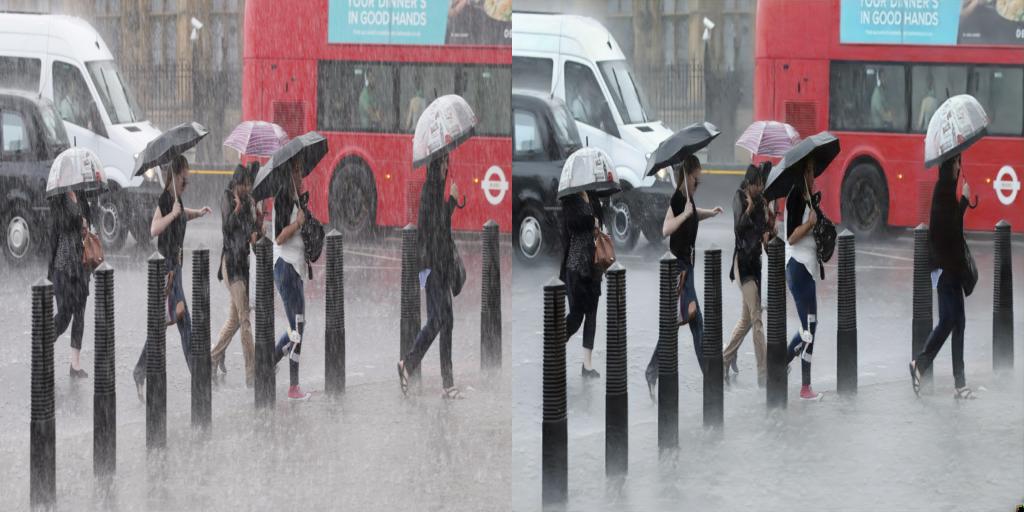}
		\captionsetup{labelformat=empty}
		%\captionsetup{justification=centering}
		%\caption*{(a) }
		%\vskip+6pt
	\end{minipage}
	\begin{minipage}{.48\textwidth}
		\centering
		\includegraphics[width=8.3cm,height=2.8cm]{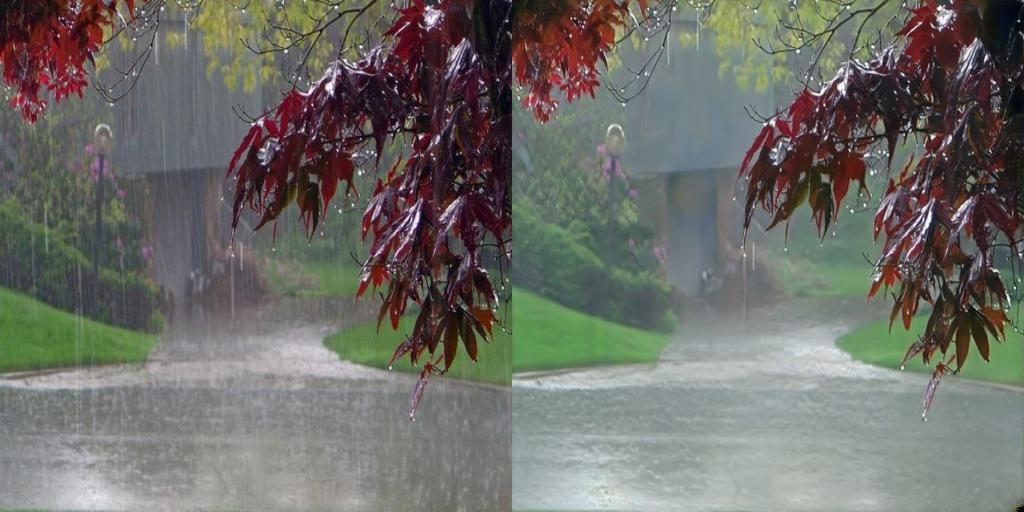}
		\captionsetup{labelformat=empty}
		%\captionsetup{justification=centering}
		%\ caption*{(b) }
		%\vskip+6pt
	\end{minipage}
	\begin{minipage}{.48\textwidth}
		\centering
		\includegraphics[width=8.3cm,height=2.8cm]{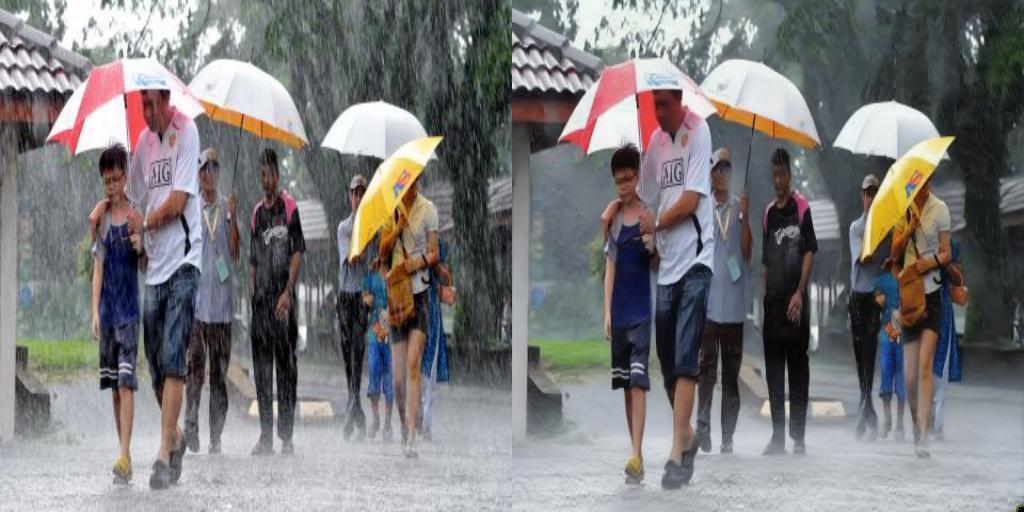}
		\captionsetup{labelformat=empty}
		%\captionsetup{justification=centering}
		%\caption*{(c) }
		%\vskip+6pt
	\end{minipage}
	\caption{Additional de-rained results using the proposed ID-CGAN method on real-world dataset. \textit{Left}: Input; \textit{Right}: Derained results. }
	\label{fig:more_images}
\end{figure}

We also evaluated the performance of the proposed method and recent state-of-the-art methods on real-world rainy test images. The de-rained results for all the methods on two sample input rainy images are shown in Fig. \ref{fig:color_ALL}. For better visual comparison, we show zoomed versions of the two specific regions-of-interest below the de-rained results.  By looking at these regions-of-interest, we can clearly observe that  DSC \cite{dis_rain_2015} tends to add artifacts on the de-rained images. Even though the other methods GMM \cite{rain_2016_gmm}, CNN \cite{derain_tip17}, CCR \cite{rain_wacv2017}, DNN \cite{derain_cvpr2017} and JORDER \cite{derain_cvpr2017_multi}  are able to achieve good visual performance, rain drops are still visible in the zoomed regions-of-interest. In comparison, the proposed method is able to remove most of the rain drops while maintaining the details of the background image. One may observe that the proposed method leaves out a few rain-streaks in the output images. This is because the two image samples represent relatively difficult cases for de-raining.  However, the proposed method is able to achieve better results compared to state-of-the-art methods. Additional comparisons are provided in Fig. \ref{fig:real2}. It can be seen that the proposed method achieves better results among all the methods. In addition, more de-rained results on different rainy images, shown in Fig.~\ref{fig:more_images}, demonstrate that the proposed method successfully removes rain streaks. 

\begin{table}[b]
	\centering
	\label{ta:object1}
	%\resizebox{0.99\textwidth}{!}{%
		\begin{tabular}{cc}
			\hline
			Condition & mAP \\
			\hline\hline
			With Rain & 0.39 \\ \hline
			DDN \cite{derain_cvpr2017} & 0.65 \\ \hline
			JORDER \cite{derain_cvpr2017} & 0.63 \\ \hline
			Our De-rained & 0.69  \\ \hline
		\end{tabular} %}
	\caption{Object detection performance using Faster-RCNN on VOC 2007 dataset. }
\end{table}

\subsubsection{Evaluation on Object Detection Results}
Single image de-raining algorithms can be used as a pre-processing step to improve the performance of other high level vision tasks such as face recognition and object detection \cite{derain_tip12}. In order to demonstrate the performance improvement obtained after de-raining using the proposed IDCGAN method, we evaluated Faster-RCNN \cite{faster_rcnn} on VOC 2007 dataset \cite{pascal}. First, the VOC 2007 dataset is artificially degraded with rain streaks similar to  Section IV A. Due to the degradations, object detection performance using Faster-RCNN results in poor performance. Next, the degraded images are processed by ID-CGAN method to remove the rain streaks and the de-rained images are fed to the Faster-RCNN method. We present the mean average precision (mAP) for the entire VOC dataset  in Table IV. It may be noted that Faster-RCNN on degraded images results in a low average precision, however, the performance is boosted by 78\% when the images undergo de-raining using the proposed ID-CGAN method.  

\begin{figure*}[htp!]
	\centering
	\begin{minipage}{.49\textwidth}
		\centering
		\includegraphics[width=8.8cm,height=3.5cm]{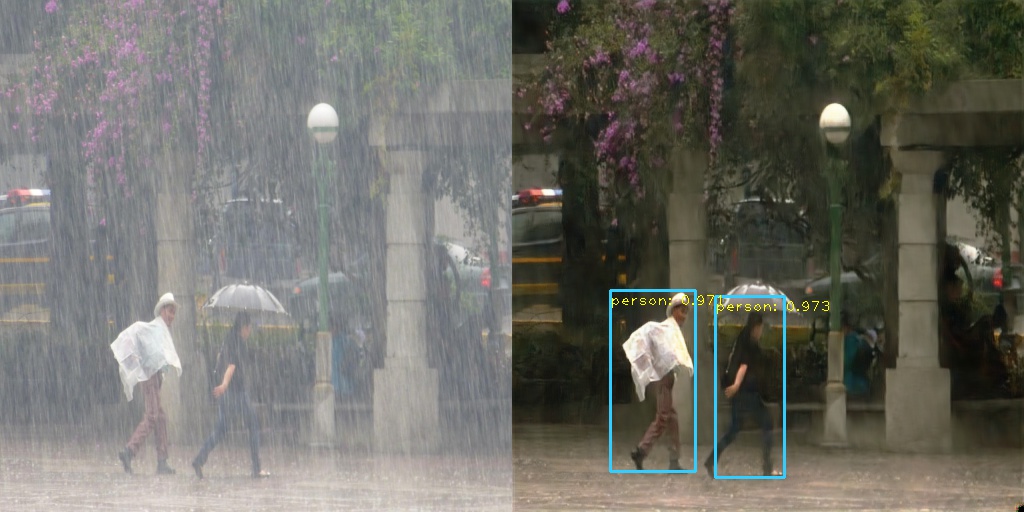}
		\captionsetup{labelformat=empty}
		\captionsetup{justification=centering}
		\caption*{(a)}
		%\vskip+6pt
	\end{minipage}
	\begin{minipage}{.49\textwidth}
		\centering
		\includegraphics[width=8.8cm,height=3.5cm]{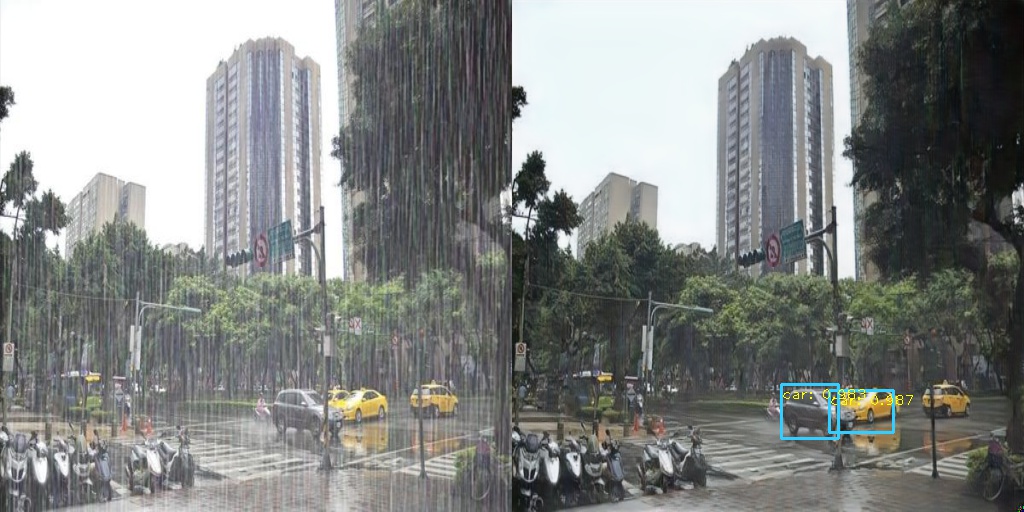}
		\captionsetup{labelformat=empty}
		\captionsetup{justification=centering}
		\caption*{(b) }
		%\vskip+6pt
	\end{minipage}	 
	\caption{Real-world examples of object detection (Faster-RCNN \cite{faster_rcnn}) improvements obtained by the proposed ID-CGAN. \emph{Left}: Detection results on rainy images; \emph{Right}: Detection results on de-rained images. The detection performance is boosted when ID-CGAN is used as a pre-processing step. } 
	\label{fig:detection_per}
\end{figure*}

Sample detection results for Faster-RCNN on real-world rainy and de-rained images are shown in Fig. \ref{fig:detection_per}. The degradations result in total failure of Faster-RCNN on these images, however, after being processed by ID-CGAN, the same detection method is able successfully detect different objects in the scene.

\begin{table*}[ht!]
	\centering
	\label{ta:time_eff}
	\resizebox{0.9\textwidth}{!}{%
		\begin{tabular}{cccccccccc}
			\hline
			& SPM \cite{derain_tip12} & PRM \cite{derain_lowrank} & DSC \cite{dis_rain_2015} & CNN (GPU) \cite{derain_tip17}  & GMM \cite{rain_2016_gmm} & CCR \cite{rain_wacv2017} & DDN (GPU) \cite{derain_cvpr2017} & JORDER (GPU) \cite{derain_cvpr2017_multi} & ID-CGAN (GPU) \\ \hline\hline
			250X250 & 400.5s & 40.5s & 1.3s & 54.9s & 169.6s & 150.2s & \textbf{0.2s} & 0.4s & \textbf{0.2s}  \\ \hline
			500X500 & 1455.2s & 140.3s & 2.8s & 189.3 & 674.8  & 600.6s & \textbf{0.3s} & 1.4s & \textbf{0.3s} \\ \hline
		\end{tabular}}
	\caption{Time complexity (in seconds) for different methods. }
	\end{table*}

\subsubsection{Computation times} 
Table \ref{ta:time_eff} compares the running time of several state-of-the-art methods. All baseline methods are implemented using MATLAB or MATLAB wrapper. Our method is implemented in Torch. It can be observed that all GPU-based CNN methods \cite{derain_tip17,derain_cvpr2017,derain_cvpr2017_multi} are computationally more efficient. The proposed ID-CGAN is able achieve the fastest time\footnote{ID-CGAN is running as fast as Fu~\emph{et al} \cite{derain_cvpr2017}.} as compared to these methods. On an average, ID-CGAN in GPU can process and image of size  500 $\times$ 500 in about 0.3s.

\section{Conclusion}\label{sec:con}
In this paper, we proposed a conditional GAN-based algorithm for the removal of rain streaks form a single image. In comparison to the existing approaches which attempt to solve the de-raining problem in an image decomposition framework by using prior information, we investigated the use of generative modeling for synthesizing de-rained image from a given input rainy image. For improved stability in training and reducing artifacts introduced by GANs in the output images, we proposed the use of a new refined loss function in the GAN optimization framework. In addition, a multi-scale discriminator is proposed to leverage features from different scales to determine whether the de-rained image is real or fake. Extensive experiments are conducted on synthetic and real-world dataset to evaluate the performance of the proposed method. Comparison with several recent methods demonstrates that our approach achieves significant improvements in terms of different metrics. Moreover, a detailed ablation study is conducted to clearly illustrate improvements obtained due to different modules in the proposed method. Furthermore, experimental results evaluated on objection detection using Faster-RCNN demonstrated significant improvements in detection performance when ID-CGAN method is used as a pre-processing step. 

%In spite of the superior performance achieved by the proposed method, it still suffers from a few drawbacks such as it fails to remove the white-round rain particles.  In the future, we aim to build upon the conditional GAN framework to overcome these drawback and investigate the possibility of using similar structures for solving related problems.

\section*{Acknowledgment} This work was supported by an ARO grant W911NF-16-1-0126.
\bibliographystyle{IEEEtran}
\bibliography{egbib}

\begin{IEEEbiography}[{\includegraphics[width=1in,height=1.75in,clip,keepaspectratio]{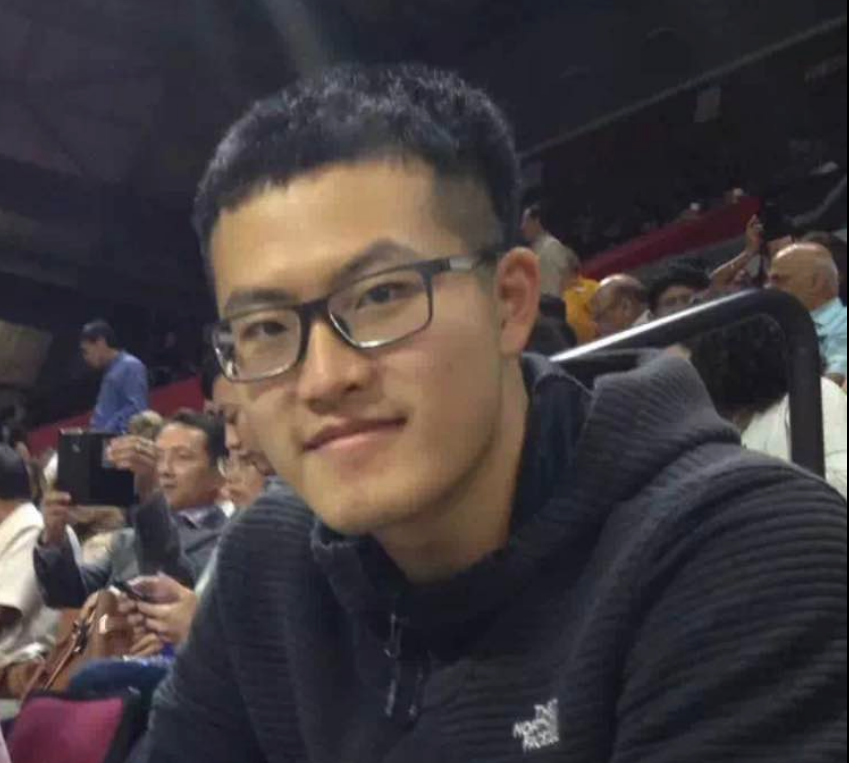}}]{He Zhang}[S'14] received his Ph.D. degree in Electrical and Computer Engineering from Rutgers University, NJ, USA in 2018. He is currently a Research Scientist in Adobe, CA.   His research interests include image restoration, image compositing, federate learning,  generative adversarial network, deep learning and sparse and low-rank representation.   
\end{IEEEbiography}
\begin{IEEEbiography}[{\includegraphics[width=1in,height=1.45in,clip,keepaspectratio]{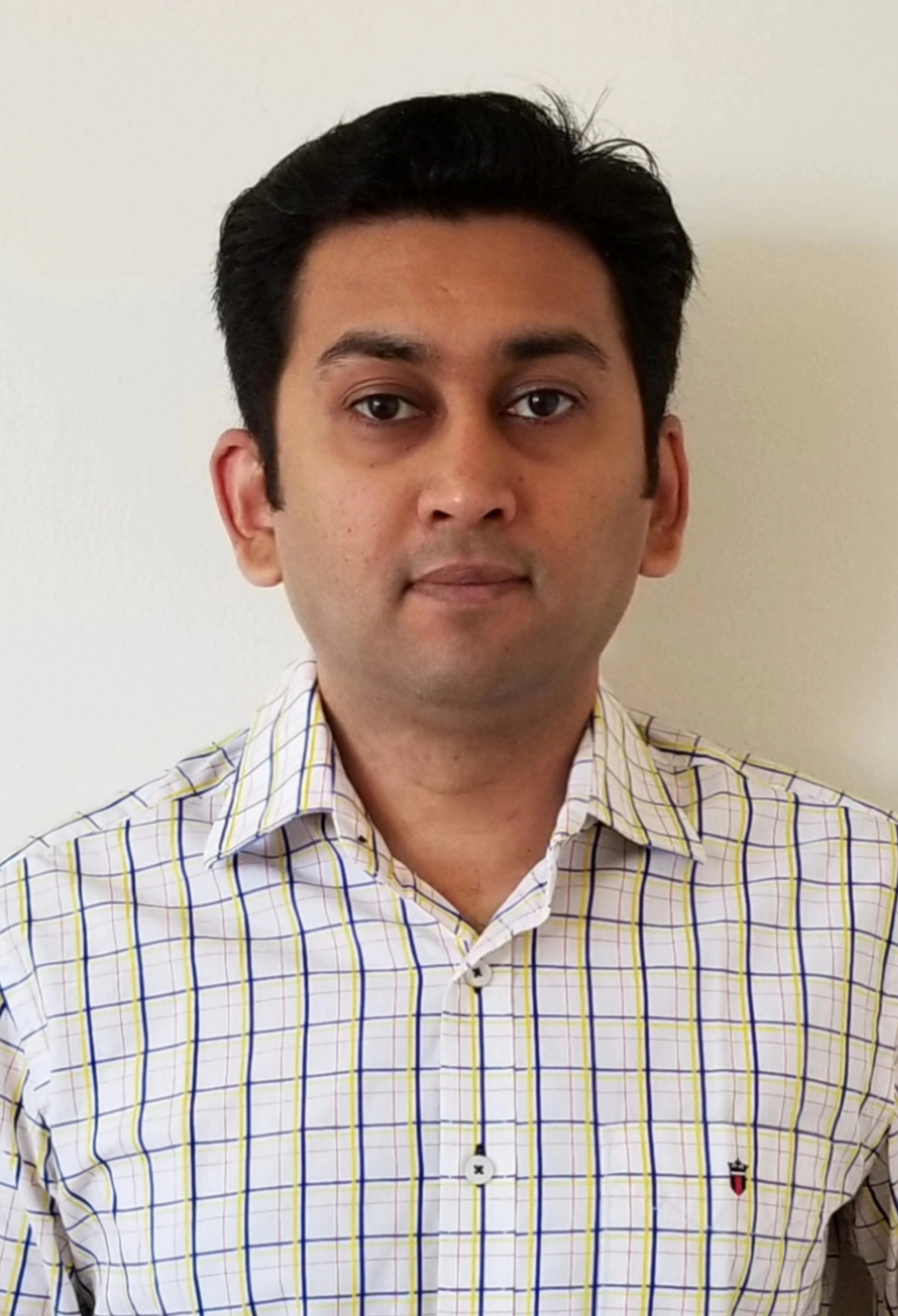}}]{Vishwanath Sindagi }[S'16] is a PhD student in the Dept. Of Electrical \& Computer Engineering at The Johns Hopkins University.  Prior to joining Johns Hopkins, he worked for Samsung R\&D Institute-Bangalore. He graduated from IIIT-Bangalore with a Master's degree in Information Technology. His research interests include deep learning based crowd analytics, object detection, applications of generative modeling, domain adaptation and low-level vision.
	Attachments area
	
\end{IEEEbiography}
\begin{IEEEbiography}[{\includegraphics[width=1in,height=1.25in,clip,keepaspectratio]{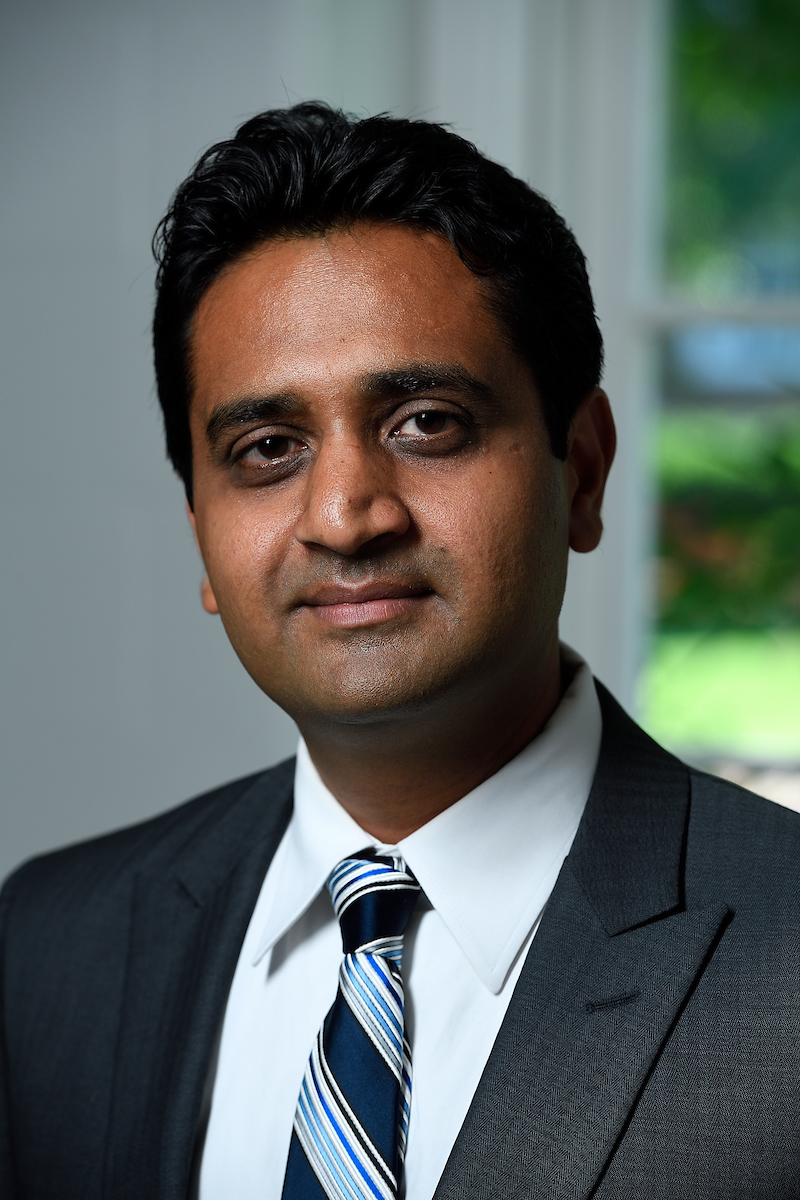}}]{Vishal M. Patel}[SM’15] is an Assistant Professor in the Department of Electrical and Computer Engineering (ECE) at Johns Hopkins University. Prior to joining Hopkins, he was an A. Walter Tyson Assistant Professor in the Department of ECE at Rutgers University and a member of the research faculty at the University of Maryland Institute for Advanced Computer Studies (UMIACS).  His current research interests include signal processing, computer vision, and pattern recognition with applications in biometrics and imaging.  He has received a number of awards including the 2016 ONR Young Investigator Award, the 2016 Jimmy Lin Award for Invention, A. Walter Tyson Assistant Professorship Award, Best Paper Award at IEEE AVSS 2017, Best Paper Award at IEEE BTAS 2015, Honorable Mention Paper Award at IAPR ICB 2018, two Best Student Paper Awards at IAPR ICPR 2018, and Best Poster Awards at BTAS 2015 and 2016. He is an Associate Editor of the IEEE Signal Processing Magazine, IEEE Biometrics Compendium, Pattern Recognition Journal, and serves on the Information Forensics and Security Technical Committee of the IEEE Signal Processing Society.  He is serving as the  Vice  President (Conferences)  of  the  IEEE  Biometrics  Council.  He is a member of Eta Kappa Nu, Pi Mu Epsilon, and Phi Beta Kappa.
\end{IEEEbiography}

\end{document}